\documentclass[sigconf, nonacm]{acmart}
\AtBeginDocument{%
  }

\usepackage{multirow}
\usepackage{subcaption}
\usepackage{pifont}
\usepackage{balance}
\begin{document}

\title{MoralCLIP: Contrastive Alignment of Vision-and-Language Representations with Moral Foundations Theory}

\author{Ana Carolina Condez}
\orcid{0000-0002-9829-0936}
\affiliation{
  \institution{NOVA LINCS, NOVA School of Science and Technology}
  \city{Caparica}
  \country{Portugal}
}
\email{a.condez@campus.fct.unl.pt}

\author{Diogo Tavares}
\orcid{0000-0002-0147-369X}
\affiliation{%
  \institution{NOVA LINCS, NOVA School of Science and Technology}
  \city{Caparica}
  \country{Portugal}}
\email{dc.tavares@campus.fct.unl.pt}

\author{Jo\~ao Magalh\~aes}
\orcid{0000-0001-6290-5719}
\affiliation{%
  \institution{NOVA LINCS, NOVA School of Science and Technology}
  \city{Caparica}
  \country{Portugal}
}
\email{jmag@fct.unl.pt}

\renewcommand{\shortauthors}{Ana Carolina Condez, Diogo Tavares, and João Magalhães}

\begin{abstract}
Recent advances in vision-language models have enabled rich semantic understanding across modalities. However, these encoding methods lack the ability to interpret or reason about the moral dimensions of content---a crucial aspect of human cognition. In this paper, we address this gap by introducing MoralCLIP, a novel embedding representation method that extends multimodal learning with explicit moral grounding based on Moral Foundations Theory (MFT). Our approach integrates visual and textual moral cues into a unified embedding space, enabling cross-modal moral alignment. MoralCLIP is grounded on the multi-label dataset Social-Moral Image Database to identify co-occurring moral foundations in visual content. For MoralCLIP training, we design a moral data augmentation strategy to scale our annotated dataset to 15,000 image-text pairs labeled with MFT-aligned dimensions.
Our results demonstrate that explicit moral supervision improves both unimodal and multimodal understanding of moral content, establishing a foundation for morally-aware AI systems capable of recognizing and aligning with human moral values.\footnotemark
\end{abstract}


\begin{teaserfigure}
  \includegraphics[width=\textwidth]{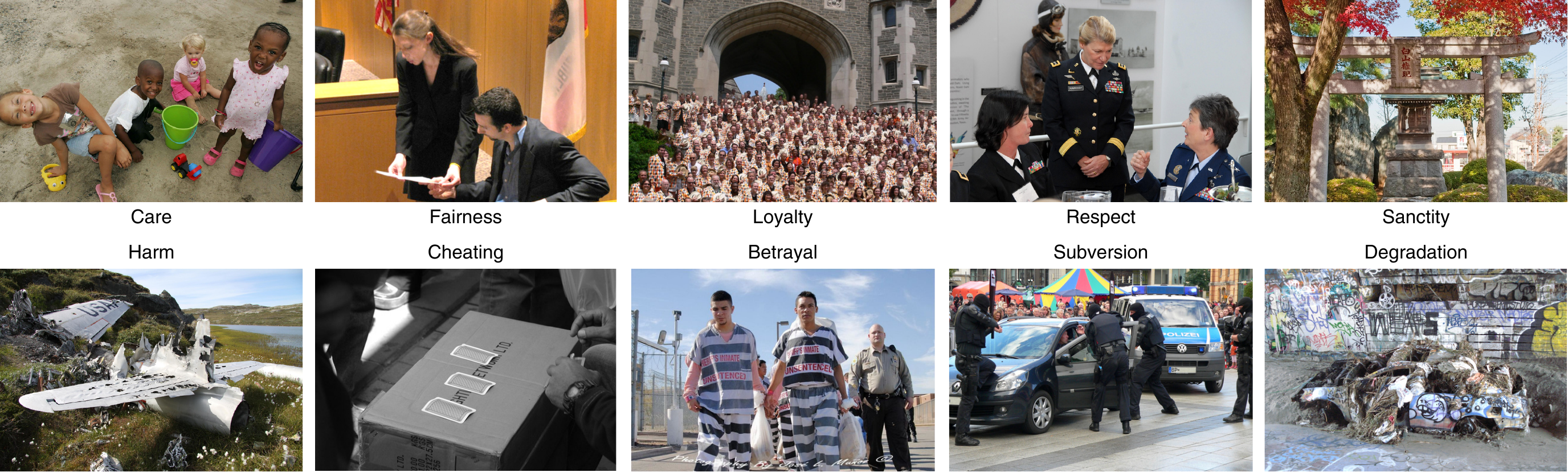}
  \caption{Visual depictions of fundamental moral dimensions as defined by Moral Foundations Theory (MFT)~\cite{MFT}. Images were extracted from the SMID~\cite{smid} dataset, which was annotated at large-scale by humans and validated by experts. MoralCLIP successfully creates a vision and language embedding space where these dimensions are well captured.}
  \label{fig:teaser}
\end{teaserfigure}

\maketitle
\section{Introduction}
\footnotetext{This paper was published in the \textit{Proceedings of the 33rd ACM International Conference on Multimedia (MM ’25)}, 
Dublin, Ireland. Association for Computing Machinery, New York, NY, USA, 12399–12408. 
\href{https://doi.org/10.1145/3746027.3758166}{https://doi.org/10.1145/3746027.3758166}. 
\\[3pt]
All supplementary material and resources available at \url{https://anaacondez.github.io/moralclip/}.}
Images are among the most powerful stimuli humans encounter, often surpassing text in their ability to instantly convey meaning and evoke emotions~\cite{socialmoral,picsup}. Unlike text, which generally requires greater cognitive effort to interpret, images can evoke intuitive moral responses almost instantly~\cite{harenski,smid,picture_this}, with a single image being capable of sparking social movements, changing public opinion, and influencing moral perceptions~\cite{socialmoral,picture_this}.
Human communication naturally integrates these modalities to construct meaning~\cite{wong2019chapter1, kress2010chapter1}, easily intertwining visual and language modalities. We process the world through this multimodal lens, where text and images interact to create richer, more nuanced understandings~\cite{imgtxt_relation, cognitive_txt_img}. This is particularly evident in moral contexts where visual cues might reinforce, contradict, or complicate textual narratives to create more powerful ethical impressions than either medium alone~\cite{imgtxt_relation, smid, maffs, picture_this, tamborini_2017}. Recent advances in artificial intelligence have begun to reflect this multimodal integration, with vision-language models like CLIP~\cite{CLIP} and SigLIP~\cite{siglip, siglip2} bridging the gap between visual and textual information. Although these models excel at semantic understanding across modalities, they have not been designed to model or interpret the moral dimensions of content. 

\begin{table}[t]
\centering
\caption{The fundamental moral foundations defined by Moral Foundations Theory (MFT)~\cite{MFT}.}
\label{tab:moral_dimensions}
\resizebox{\columnwidth}{!}{%
\begin{tabular}{@{}lp{6cm}@{}}
\toprule
\textbf{Moral Foundation} & \textbf{Description} \\ \midrule
Care/Harm                 & Intuitions about preventing emotional and physical suffering.                                                          \\[0.8ex]
Fairness/Cheating         & Intuitions regarding justice, rights, and equitable treatment in social interactions.                                      \\[0.8ex]
Loyalty/Betrayal          & Intuitions related to loyalty, obligations of group membership (in-groups), and vigilance against threats from external groups (out-groups). \\[0.8ex]
Respect/Subversion        & Intuitions about respecting and obeying higher authorities.                                                                                \\[0.8ex]
Sanctity/Degradation      & Intuitions concerned with bodily and spiritual cleanliness, as well as protection from contamination or impurity.                          \\
\bottomrule
\end{tabular}%
}
\end{table}
To analyze and understand different expressions of morality, Moral Foundations Theory (MFT)~\cite{MFT} provides a widely adopted framework for evaluating moral judgments across cultures. MFT postulates that moral reasoning is shaped by five innate moral foundations, which are thought to be universal across human societies. While this theory has been extensively applied to text analysis through both lexicon-based approaches~\cite{eMFD, MFD, jmfd2024,mft_dt} and language models~\cite{DAMF, MoralBERT, ME2BERT, mformer, park_2024}, recent attempts to extend it to visual analysis have been limited in scope and depth. Thus, a truly integrated multimodal approach to moral foundation analysis remains underexplored.

Current methods for moral analysis operate in isolated modalities, with text-only models analyzing written content and image-based approaches relying primarily on textual supervision~\cite{classifier2022}. This unimodal framing is reflected not only in model architectures but also at the data level, given most datasets constructed for moral analysis---such as Moral Foundations Twitter Corpus (MFTC)~\cite{MFT_twitter}, Moral Foundations Reddit Corpus (MFRC)~\cite{MFT_Reddit}, Moral Events~\cite{zhang_moka_2024}, and E2MoCase~\cite{greco2024e2mocase}---are exclusively textual, limiting the development of models capable of integrating multimodal moral content. While a few datasets have begun to explore morality in visual content, such as the Social-Moral Image Database~\cite{smid} and the Moral Affective Film Set~\cite{maffs}, these again represent single-modality efforts, offering no way to study how text and images jointly influence moral interpretation. Simultaneously, most vision-based datasets related to values or judgment concentrate on general notions of safety, appropriateness, and societal biases~\cite{VIVA, mmbias, mmsafetybench, dear}, rather than pluralistic morality grounded in frameworks like MFT. Furthermore, existing moral analysis methods often reduce morality to a binary concept, without considering the interplay between the different moral foundations~\cite{classifier2022, MoralBERT, mformer}, effectively overlooking the pluralistic nature of moral reasoning.
To address these challenges, we propose MoralCLIP, a framework that extends vision-language models to capture moral dimensions across both visual and textual modalities. To the best of our knowledge, this is the first attempt to leverage MFT to create a morally-grounded multimodal embedding space capable of identifying moral content within each modality. Specifically, our contributions are:
\begin{itemize}
    \item \textbf{MoralCLIP:} Building on the MFT foundations, we introduce \textit{MoralCLIP}, a novel extension of the CLIP framework that incorporates moral supervision into the contrastive learning objective. Rather than aligning visual and textual inputs based solely on semantic similarity, \textit{MoralCLIP} aligns them based on shared moral meaning, enabling it to \textit{identify similar moral foundations across different modalities even when the semantic content differs significantly}. 
    This approach creates a joint embedding space where moral dimensions take precedence over purely semantic relationships.
    \item \textbf{Morally-Grounded Multimodal Data Augmentation:} Leveraging the expert annotated dataset SMID~\cite{smid}, we construct a dataset of 15,000 image-text pairs annotated with MFT-aligned moral labels. 
    The data augmentation process is achieved with a high-precision moral image classifier, \textit{Visual Moral Compass}, and the generation of short, descriptive captions. 
    We apply this process to the ImageNet~\cite{imagenet2014} and LAION-400M~\cite{LAION_400M} datasets, while retaining SMID's~\cite{smid} expert labels, resulting in the first dataset that explicitly connects moral foundations across both modalities.
\end{itemize}

While current V\&L models, e.g. CLIP~\cite{CLIP} and SigLIP~\cite{siglip}, excel at semantic understanding, experimental results demonstrate that MoralCLIP successfully captures moral dimensions across modalities, representing a significant first step towards responsible, ethically-aligned multimodal AI that understands not just \textit{what} we communicate, but the \textit{values} behind it.

\section{Related Work}
\noindent
\textbf{Moral Foundations Theory.} \hspace{2mm}
Moral Foundations Theory (MFT) is a moral psychology framework for understanding how moral foundations shape moral reasoning across diverse cultures. While universal, MFT emphasizes that their specific expressions are deeply influenced by socio-cultural contexts and individual experiences~\cite{MFT, narrative}. At its core, MFT posits that moral judgments and decisions arise from emotional, innate evaluations known as \textit{moral intuitions}~\cite{MFT, tamborini_2017}. These intuitions allow individuals to make rapid, unconscious moral assessments---such as approving or disapproving of an action---based on their moral values.
Specifically, MFT identifies five distinct, yet interconnected moral foundations: \textit{Care}, \textit{Fairness}, \textit{In-group} (or \textit{Loyalty}), \textit{Authority} (or \textit{Respect}), and \textit{Purity} (or \textit{Sanctity}). Each of these foundations is structured as a duality, encompassing both a virtue, representing morally positive behavior, and a vice, representing morally reprehensible actions~\cite{MFT} (Table~\ref{tab:moral_dimensions}). In this study, we leverage these five moral foundations as the theoretical backbone of our morally aligned embedding space.

\noindent \textbf{Morality Encoded in Text.} \hspace{2mm}
Several works have analyzed morality encoded in text using MFT, primarily in social media~\cite{MoralBERT,Japanese_twitter,hoover2018}, news articles~\cite{tamborini_2017, eMFD}, and politics~\cite{ballot}.
Early lexicon-based approaches neglected contextual nuances~\cite{MFD, weber_2018, moral_rhetoric}, amplified annotation biases~\cite{dicts_dists}, and suffered from limited adaptability to multilingual contexts and language evolution~\cite{jmfd2024, mft_dt, park_2024}. While subsequent probabilistic and crowd-sourced lexicons improved precision~\cite{eMFD}, these methods remain rigid and difficult to scale.

In response to these challenges, recent approaches employed embedding-based methods to model moral similarity, capturing the pluralistic nature of morality~\cite{park_2024}, while others have framed moral inference explicitly as a classification task~\cite{MoralBERT, DAMF, mformer}. For domain invariance,~\citet{DAMF} and~\citet{MoralBERT} use adversarially trained BERT~\cite{BERT} models, while~\citet{mformer} fine-tunes RoBERTa~\cite{roberta} for foundation-level classification. However, these approaches rely on binary classification, which fundamentally contradicts one of the core principles of MFT: multiple foundations often co-occur and influence one another. Capturing this complexity requires multi-label models that go beyond isolated, foundation-by-foundation inference.

\noindent \textbf{Morality Encoded in Images.} \hspace{2mm} 
Few efforts have extended moral analysis to visual content~\cite{classifier2022, m3oralbench_2024}, where moral cues are conveyed through expressions, actions, and spatial relationships rather than explicit language~\cite{picture_this, m3oralbench_2024}. \citet{classifier2022} perform zero-shot binary moral classification using CLIP embeddings trained on text-only data~\cite{ethics_data} and apply the resulting classifier to image embeddings. However, this method cannot capture MFT's dimensional complexity or implicit visual cues. This highlights the need for approaches that can capture MFT's dimensional complexity through direct visual supervision and true multimodal reasoning.

\noindent \textbf{Vision-Language Pretrained Models.} \hspace{2mm}
Large-scale pretraining on vision–language data has driven progress in cross-modal representation learning. Models such as CLIP~\cite{CLIP}, ALIGN~\cite{ALIGN}, and SigLIP~\cite{siglip,siglip2} learn aligned image-text embeddings via dual-encoder architectures, enabling strong performance in zero-shot image classification and image-text retrieval. These models process visual and textual inputs independently, projecting them into a shared embedding space, and can be fine-tuned for various downstream tasks requiring cross-modal understanding~\cite{siglip2, anomalyclip, pubmedclip}. Recent work has explored modifying these embedding spaces for safety purposes. Safe-CLIP~\cite{safeclip} fine-tunes CLIP to reduce sensitivity to NSFW content, redirecting inappropriate inputs to safer embedding regions while preserving CLIP's embedding space structure. However, this approach targets safety filtering rather than moral understanding. In contrast, we leverage the multimodal capabilities of vision-language models to design a morality-aware embedding space that captures the moral dimensions embedded in visual and textual content. Such a space would enable moral analysis grounded in both textual and visual modalities.

\section{A Multimodal Moral Embedding Space}
In this section, we propose MoralCLIP, a model that extends the standard CLIP architecture to jointly align image and text representations while incorporating moral information. We consider a morally annotated dataset $ \mathcal{D}=\{(v_1,t_1,m_1), \ldots, (v_i,t_i,m_i), \ldots\}$, where each tuple is composed of an image $v_i$, the corresponding text caption $t_i$, and a set of moral labels $m_i$ indicating \textit{virtue}, \textit{neutral}, or \textit{vice} for each of the five moral foundations, Table~\ref{tab:moral_dimensions}. 

\subsection{CLIP with Implicit Moral}\label{subsec:implicit}
Building upon a pretrained CLIP model, our approach integrates MFT labels into the learning objective, encouraging the model to learn morally grounded embeddings across modalities.  
Initially, we trained CLIP in an \textit{implicit} setting, using its original contrastive loss objective, with the augmented moral dataset (Section~\ref{subsec:train_dataset}). 
We introduce moral information through morally charged images and their corresponding captions, i.e. image-text pairs from dataset $\mathcal{D}$, excluding the moral annotations completely. No architectural or loss modifications are made. This allows us to assess whether CLIP can passively acquire moral information without being guided by the labels.

\subsection{MoralCLIP}\label{subsec:explicit}
In contrast to the previous approach, MoralCLIP \textit{explicitly} encodes moral information through a dedicated loss component. Particularly, we extend CLIP's training objective to include a moral alignment term that encourages embeddings to capture MFT-based relationships. The total loss becomes a weighted combination of CLIP's original contrastive loss and our moral loss:
\begin{equation}
\mathcal{L}_{\text{Total}} = \mathcal{L}_{\text{CLIP}} + \lambda \cdot \mathcal{L}_{\text{Moral}},
\end{equation}
where $\lambda$ controls the influence of the moral alignment component. The moral term $\mathcal{L}_{\text{Moral}}$ penalizes discrepancies between the moral similarity of all sample pairs and their semantic similarity in the joint embedding space, computed as the mean squared error:
\begin{equation}
    \mathcal{L}_{\text{Moral}} = \frac{1}{N}\sum_{i,j \in B, i \neq j}\left(\text{sim}(v_i^e, t_j^e) - \text{sim}_{\text{Moral}}(M_{v_i}, M_{t_j})\right)^2
\end{equation}
where $N$ represents the total number of non-diagonal pairs in the batch, $\text{sim}(v_i^e, t_j^e) = \langle \hat{v}_i^e, \hat{t}_j^e \rangle/\tau$ is the scaled cosine similarity between normalized image ($\hat{v}_i^e$) and text ($\hat{t}_j^e$) embeddings. Following CLIP's approach~\cite{CLIP}, we apply temperature scaling $\tau$ to embedding similarity, encouraging more discriminative representations. Note that our moral loss does not optimize semantic similarity independently. Rather, it constrains the embedding space so that semantic relationships align with the moral similarity patterns defined by the moral labels. The moral similarity  $\text{sim}_{\text{Moral}}(M_{v_i}, M_{t_j})$ is computed as the scaled Jaccard Index between MFT labels of the image ($M_{v_i}$) and text ($M_{t_j}$) embeddings:
\begin{equation}
\text{sim}_{\text{Moral}}(M_{v_i}, M_{t_j}) = 2\frac{|M_{v_i} \cap M_{t_j}|}{|M_{v_i} \cup M_{t_j}|} - 1
\end{equation}
This formulation preserves CLIP's semantic similarity loss while adding the constraint that semantic similarity should align with moral overlap between samples. This is particularly suitable for our multi-label setting, where samples can be associated with multiple moral foundations simultaneously. To avoid trivial self-similarity effects, we exclude diagonal terms from the loss computation. Full training configuration details are provided in Appendix~\ref{app:training_parameters}.

\subsection{Morally-Grounded Data Augmentation}
\label{sec:visual_moral_compass}
The SMID dataset contains 2,941 images annotated along the five MFT foundations, validated by experts and rated by 2,716 individuals who provided a total of 820,525 ratings~\cite{smid}. While the data quality is high, SMID's size represents a bottleneck for large-scale contrastive training. Thus, we use it to train the \emph{Visual Moral Compass}, a multi-label classifier that predicts the presence of the five moral foundations in terms of \textit{virtue}, \textit{vice} or \textit{neutral}, constituting an essential element in our workflow to obtain a broader training dataset.

\noindent\textbf{Visual Moral Classifier.} \hspace{2mm}
The \emph{Visual Moral Compass} is a high-quality image moral classifier, built on a fine-tuned CLIP (ViT-B/16) vision encoder, followed by five independent classifier heads---one for each moral foundation (architecture and training details in Appendix~\ref{app:vmc}). Only the final encoder layer and classifier heads are trained. Each head outputs a probability distribution over three classes: \textit{virtue}, \textit{vice}, or \textit{neutral}, enabling multi-label classification that reflects the pluralistic nature of moral judgment~\cite{MFT}. The classifier is trained on a preprocessed version of SMID consisting of 2,401 entries, where each image is labeled according to its moral dimension within a given foundation. Detailed preprocessing steps are described in Appendix~\ref{app:vmc_smid_preprocess}.

\noindent \textbf{Moral Data Labeling.} \hspace{2mm}
~\label{subsec:train_dataset}
To create a morally grounded multimodal dataset, we used the \emph{Visual Moral Compass} to annotate large-scale, unlabeled image datasets with MFT-aligned moral labels. 
We applied the classifier to two source datasets: the ImageNet validation set~\cite{imagenet2014} and a 10M subset of LAION-400M~\cite{LAION_400M}. 
ImageNet, originally developed as an object classification benchmark, contains images organized into 1000 categories, and has previously been shown to include morally relevant content~\cite{classifier2022}. 

\noindent \textbf{Moral Captions.} \hspace{2mm}
When generating captions for morally negative images, most models tend to refuse responses or sanitize descriptions~\cite{paligemma, instructblip}.
To mitigate this limitation, we used the MoonDream2B captioning model~\cite{moondream2b}\footnote{Model available at \href{https://huggingface.co/vikhyatk/moondream2}{HuggingFace}, version from 26-08-2024.} which can generate accurate depictions of morally negative scenes---an essential property for our setting.
Ten captions per image were generated for the ImageNet and SMID datasets, and five captions per image for the LAION dataset due to its larger size. The captions are concise descriptions of the images, making them well-suited for CLIP's context window.

\begin{figure}[t]
    \centering
    \begin{subfigure}{\columnwidth}
        \centering
        \includegraphics[width=\textwidth]{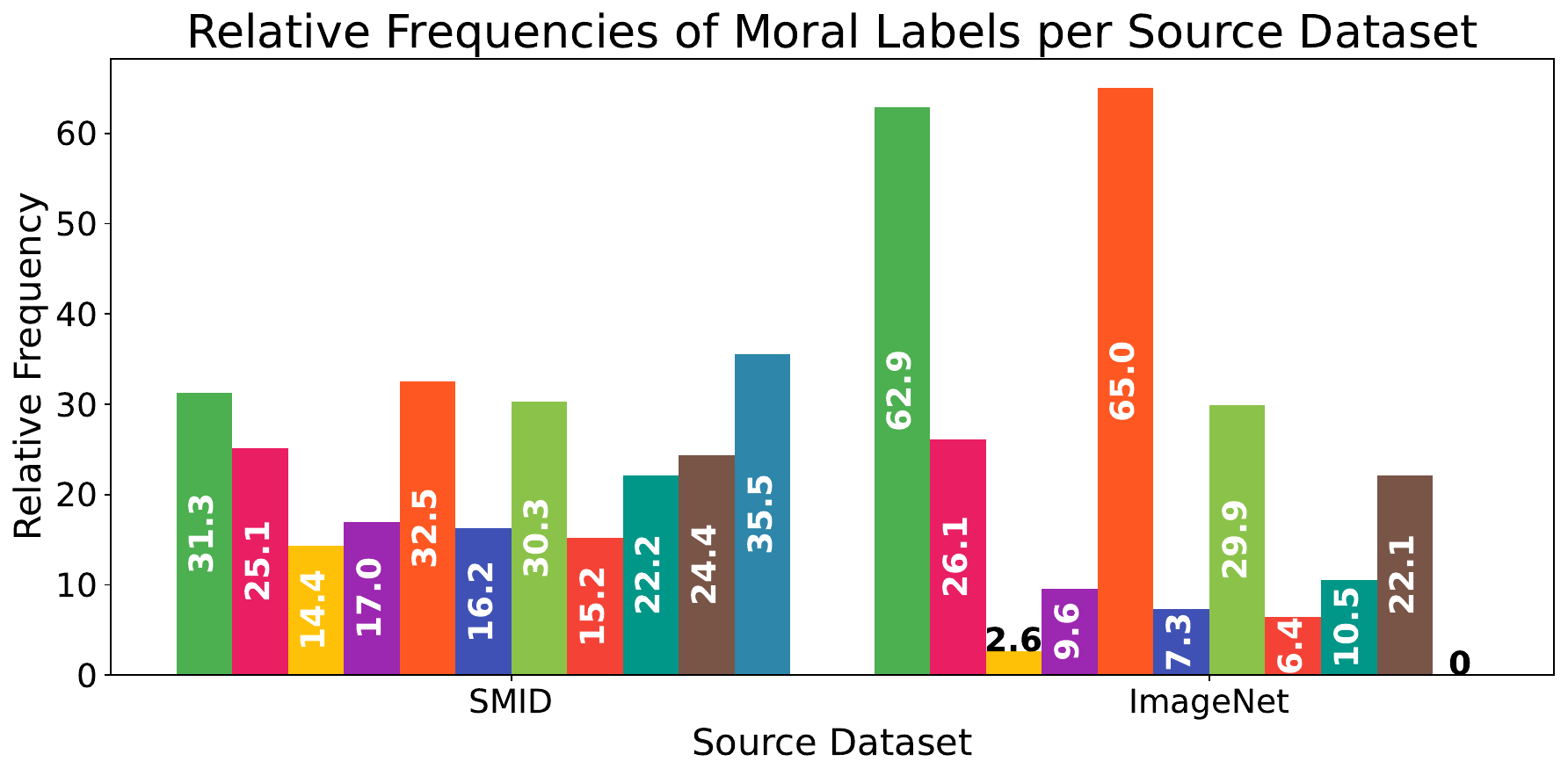}
    \end{subfigure}
    \begin{subfigure}{\columnwidth}
        \centering
        \includegraphics[width=\textwidth]{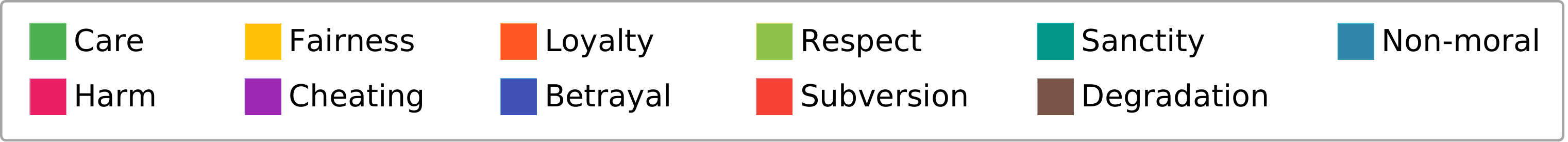}
        \label{fig:train_dataset_legend}
    \end{subfigure}%
    \caption{Dataset distribution showing moral label frequencies across SMID (2,401 preprocessed samples) and ImageNet (10,602 samples) from our 15,000-sample training set. LAION samples (1,997) are omitted as they contain exclusively neutral moral content.}
    \label{fig:train_dataset}
\end{figure}

This process yielded a training set of 15,000 images, with their distribution across data sources showcased in Figure~\ref{fig:train_dataset}. Due to the different characteristics of our source datasets, we retained morally relevant ImageNet samples and neutral LAION examples---ImageNet contains more morally relevant scenes while LAION's morally charged content was both rare and typically benign. This was essential for achieving balanced moral representation and avoiding severe class imbalance. Both ImageNet and LAION have been extensively filtered and curated for their original purposes, which naturally reduces the prevalence of explicit moral content compared to specialized datasets like SMID.

\noindent{\textbf{Agreement.}} \hspace{2mm}
This pipeline resulted in a weakly labeled dataset of paired images and captions, each annotated with predicted moral foundation labels. To validate our automated labeling approach, we conducted a human evaluation (see Section~\ref{subsec:vmc_agree} for summary results; detailed annotation procedure and interface are provided in Appendix~\ref{app:human_annotations}, and full classifier metrics in Appendix~\ref{app:vmc_performance}). 
.

\begin{table*}[htbp]
\caption{Performance comparison of MoralCLIP variants across modalities on the SMID test subset. We report Mean Average Precision (MAP) to measure retrieval performance of morally similar content, Discriminative Power (DP) to quantify separation between moral categories, and Silhouette Score to assess embedding separation quality. For cross-modal evaluation we use  MAP for both image-to-text (I2T) and text-to-image (T2I) retrieval. Best performing models are in bold. M and S denote Mild and Strong MFT mixing strategies, respectively. All metrics include standard errors computed via bootstrap resampling ($n=1000$).}
\label{tab:moral_clip_metrics}
\resizebox{\textwidth}{!}{%
\begin{tabular}{@{}l l c c c c c c c c l@{}}
\toprule
\multirow{2}{*}{\textbf{Model type}} 
  & \multicolumn{1}{l}{\multirow{2}{*}{\textbf{Variant}}} 
     & \multicolumn{3}{c}{\textbf{Image}} 
        & \multicolumn{3}{c}{\textbf{Text}} 
           & \multicolumn{2}{c}{\textbf{Cross-modal}} \\
\cmidrule(lr){3-5} \cmidrule(lr){6-8} \cmidrule(lr){9-10}
  &  
  & \textbf{MAP} & \textbf{DP} & \textbf{Silhouette} 
       & \textbf{MAP} & \textbf{DP} & \textbf{Silhouette} 
          & \textbf{MAP-I2T} & \textbf{MAP-T2I} \\
\midrule               
CLIP~\cite{CLIP} &      --            & $42.06 \pm 0.92$                                & \multicolumn{1}{c}{$1.051 \pm 0.009$}               & \multicolumn{1}{c}{$0.013 \pm 0.005$}              & $38.73 \pm 0.77$                                & \multicolumn{1}{c}{$1.035 \pm 0.013$}              & \multicolumn{1}{c}{$-0.000 \pm 0.004$}             & $39.82 \pm 0.81$                                & \multicolumn{1}{c}{$39.85 \pm 0.79$}               \\
Safe-CLIP~\cite{safeclip} &        --          & $43.91 \pm 1.04$                                & \multicolumn{1}{c}{$1.025 \pm 0.005$}               & \multicolumn{1}{c}{$0.005 \pm 0.006$}              & $39.19 \pm 0.83$                                & \multicolumn{1}{c}{$1.053 \pm 0.022$}              & \multicolumn{1}{c}{$0.000 \pm 0.006$}             & $40.51 \pm 0.87$                                & \multicolumn{1}{c}{$40.97 \pm 0.89$}               \\
\midrule
\multirow{4}{*}{CLIP+Moral Images}                     & Normal                                      & $43.00 \pm 0.97$                                & \multicolumn{1}{c}{$1.141 \pm 0.015$}               & \multicolumn{1}{c}{$0.013 \pm 0.005$}           & $41.07 \pm 0.97$                                & \multicolumn{1}{c}{\textbf{1.148 $\pm$ 0.024}}              & \multicolumn{1}{c}{$0.006 \pm 0.005$}             & $41.53 \pm 0.93$                                & \multicolumn{1}{c}{$41.26 \pm 0.94$}               \\
                                                      & Augmented                                   & $42.39 \pm 0.97$                                & \multicolumn{1}{c}{$1.132 \pm 0.013$}               & \multicolumn{1}{c}{$0.012 \pm 0.005$}           & $41.00 \pm 0.99$                                & \multicolumn{1}{c}{$1.120 \pm 0.023$}              & \multicolumn{1}{c}{$0.007 \pm 0.005$}             & $41.29 \pm 2.20$                                & \multicolumn{1}{c}{$41.13 \pm 1.00$}               \\
                                                      & MFT Swapper (M)                             & $51.10 \pm 1.43$                                & \multicolumn{1}{c}{$1.049 \pm 0.003$}               & \multicolumn{1}{c}{$0.036 \pm 0.008$}           & $45.82 \pm 1.12$                                & \multicolumn{1}{c}{$1.014 \pm 0.002$}              & \multicolumn{1}{c}{$0.014 \pm 0.006$}             & $49.14 \pm 1.25$                                & \multicolumn{1}{c}{$50.95 \pm 1.61$}               \\
                                                      & MFT Swapper (S)                             & $53.77 \pm 1.60$                                & \multicolumn{1}{c}{$1.051 \pm 0.003$}               & \multicolumn{1}{c}{$0.045 \pm 0.009$}           & $46.34 \pm 1.11$                                & \multicolumn{1}{c}{$1.014 \pm 0.002$}              & \multicolumn{1}{c}{$0.012 \pm 0.006$}             & $50.81 \pm 1.41$                                & $50.97 \pm 2.14$                                    \\ 
\midrule
\multirow{4}{*}{%
  \begin{tabular}[c]{@{}l@{}}
    \textbf{MoralCLIP}
  \end{tabular}%
}                   & Normal ($\lambda=0.5$)          & $65.51 \pm 2.58$                                & \multicolumn{1}{c}{$1.160 \pm 0.008$}               & \multicolumn{1}{c}{$0.084 \pm 0.014$}           & $58.61 \pm 1.96$                                & \multicolumn{1}{c}{$1.071 \pm 0.006$}              & \multicolumn{1}{c}{$0.048 \pm 0.011$}             & $63.88 \pm 1.75$                                & $65.73 \pm 2.31$                                    \\
                                                      & Augmented ($\lambda=0.4$)       & \textbf{71.68 $\pm$ 2.02}                                & \multicolumn{1}{c}{1.187 $\pm$ 0.008}               & \multicolumn{1}{c}{\textbf{0.107 $\pm$ 0.016}}           & \textbf{61.77 $\pm$ 2.04}                                & \multicolumn{1}{c}{1.075 $\pm$ 0.006}              & \multicolumn{1}{c}{\textbf{0.058 $\pm$ 0.013}}             & \textbf{64.37 $\pm$ 1.76}                               & \textbf{66.83 $\pm$ 2.27}                                    \\
                                                      & MFT Swapper (M) ($\lambda=0.5$) & $68.23 \pm 2.09$                                & \multicolumn{1}{c}{\textbf{1.253 $\pm$ 0.014}}               & \multicolumn{1}{c}{$0.084 \pm 0.013$}           & $57.00 \pm 1.93$                                & \multicolumn{1}{c}{$1.042 \pm 0.004$}              & \multicolumn{1}{c}{$0.034 \pm 0.008$}             & $62.16 \pm 1.68$                                & $63.24 \pm 2.22$                                    \\
                                                      & MFT Swapper (S) ($\lambda=0.5$) & $66.13 \pm 2.14$                                & \multicolumn{1}{c}{$1.207 \pm 0.014$}               & \multicolumn{1}{c}{$0.072 \pm 0.011$}           & $56.40 \pm 1.94$                                & \multicolumn{1}{c}{$1.043 \pm 0.005$}              & \multicolumn{1}{c}{$0.032 \pm 0.008$}             & $61.02 \pm 1.72$                                & $62.31 \pm 2.18$                                    \\ 
\bottomrule
\end{tabular}%
}
\end{table*}

\section{Experimental Setup}
In this section, we describe the key details of the baselines (Section~\ref{subsec:baselines}), datasets (Section~\ref{subsec:datasets}), and metrics (Section~\ref{subsec:metrics}).

\subsection{Baselines}\label{subsec:baselines}
To systematically assess the effectiveness of the proposed methods, we trained different variants of MoralCLIP and implicitly trained CLIP models, comparing them against CLIP and Safe-CLIP~\cite{safeclip}, a model designed to filter NSFW content.

\noindent \textbf{Normal.} \hspace{2mm} Standard fine-tuning without conventional data augmentation, serving as our baseline. This variant establishes the model's base capacity to learn moral associations purely from the original dataset.

\noindent \textbf{Augmented.} \hspace{2mm} Enhances the training with standard data augmentation\footnote{Specific augmentation techniques include random horizontal flips, color jittering, and random resized crops.}. Each training sample is augmented four times, with each augmented version paired with alternative---but still semantically accurate---captions from our dataset. 
These augmentations preserve moral content while varying visual properties, testing whether visual robustness alone improves generalization in moral grounding.

\noindent \textbf{MFT Swapper.} \hspace{2mm} Building upon the Augmented variant, this version implements content mixing between samples that share moral dimensions for 75\% of the dataset.
Specifically, we randomly swap images and text descriptions between samples with matching moral labels, creating new image-text pairs while preserving their moral associations. 
This approach assesses if the model can learn moral concepts that generalize further than specific image-text instances---for example, whether it can recognize that different manifestations of \textit{Care}, such as helping the elderly, sharing food, caring for animals, represent the same underlying moral dimension.  
We explore two levels of this mixing strategy: (1) \textbf{Mild}, which limits the number of swaps to a maximum of 500 per moral dimension group to ensure uniform sampling across moral concepts; and (2) \textbf{Strong}, which allows more frequent moral labels to be sampled more often.

This experimental design allows us to systematically evaluate different approaches to improving moral understanding in multimodal models. 
Training each variant under both \textit{implicit} (Section~\ref{subsec:implicit}) and \textit{explicit} (Section~\ref{subsec:explicit}) supervision enables us to assess whether these enhancements are more effective when moral learning is guided explicitly or emerges implicitly through contrastive training. 
Both visual and text encoder models were initialized from CLIP's ViT-B/16, with full fine-tuning of all layers during the contrastive moral alignment training.

\subsection{Datasets}\label{subsec:datasets}
The total number of samples in our dataset $\mathcal{D}$ is 15,000 morally labeled image-text pairs. We used standard held-out splits for evaluation: 5\% validation and 5\% test splits from all three source datasets for MoralCLIP assessment. Both splits preserve the relative distribution of moral foundations to ensure balanced evaluation.

For model selection, we exclusively used the SMID portion of the validation set for computing metrics, as it contains the strongest moral signal with expert-curated annotations. For final evaluation, we used different test set portions depending on the analysis. Retrieval analysis used exclusively queries from the SMID test subset to leverage its reliable moral signal, while embedding visualization analyses used the complete test set to provide a comprehensive view across different data distributions.

\begin{figure*}[t]
\centering
\begin{subfigure}[t]{0.24\textwidth}
    \centering
    \includegraphics[width=\textwidth, trim=0 3 0 0, clip]{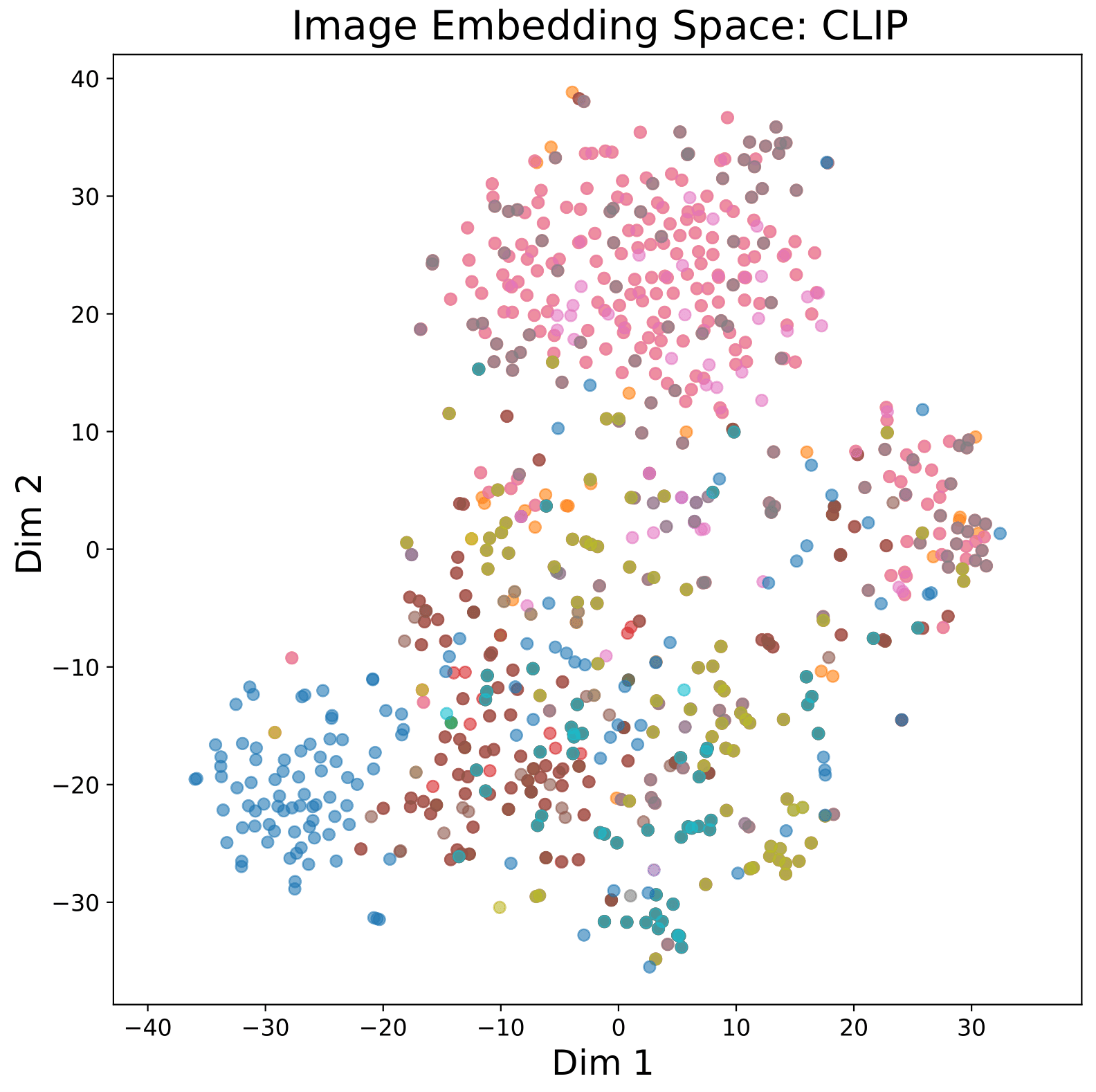}
    \label{fig:tsne_clip_image}
\end{subfigure}
\hfill
\begin{subfigure}[t]{0.24\textwidth}
    \centering
    \includegraphics[width=\textwidth]{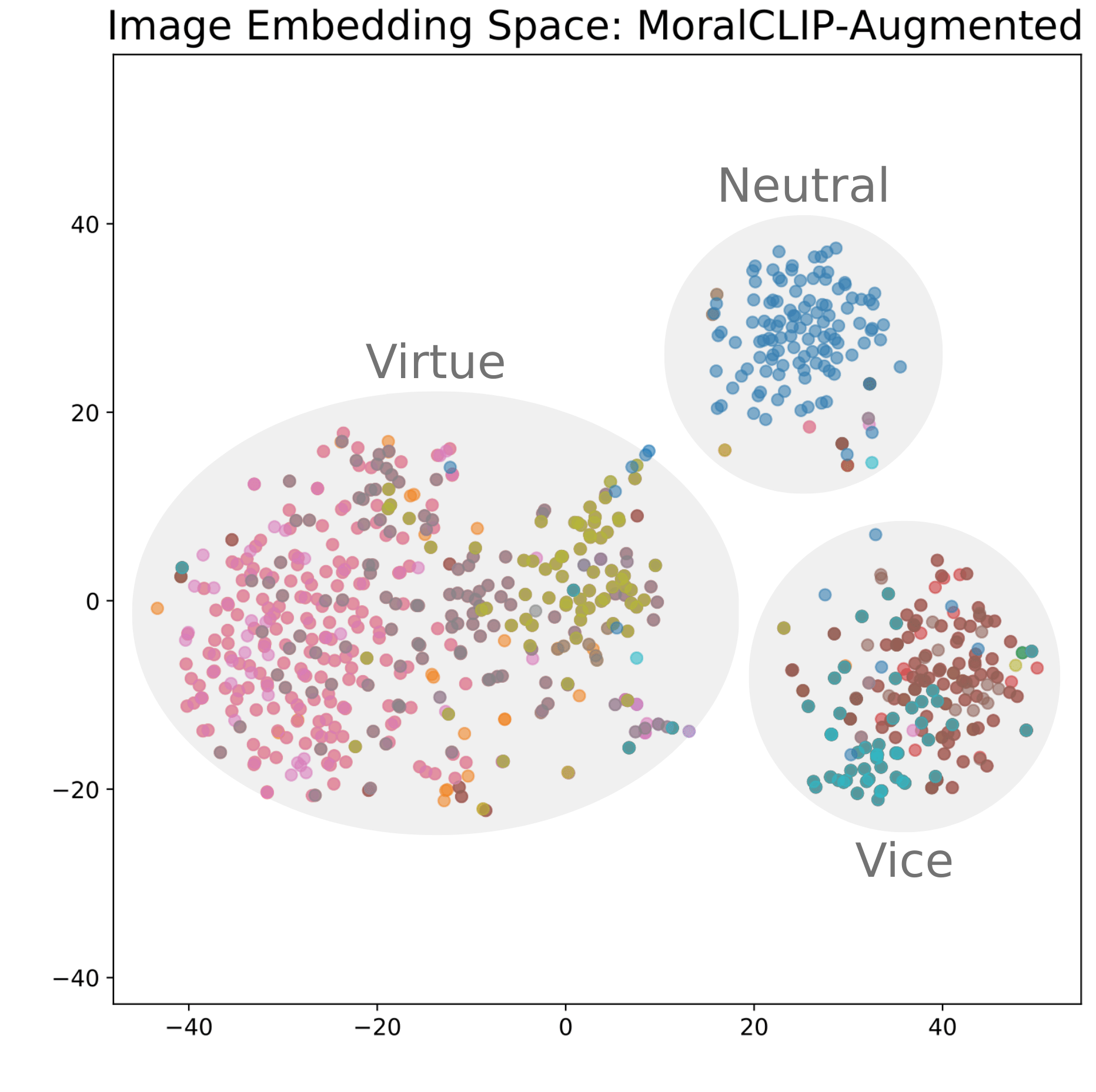}
    \label{fig:tsne_moralclip_image}
\end{subfigure}
\hfill
\begin{subfigure}[t]{0.24\textwidth}
    \centering
    \includegraphics[width=\textwidth]{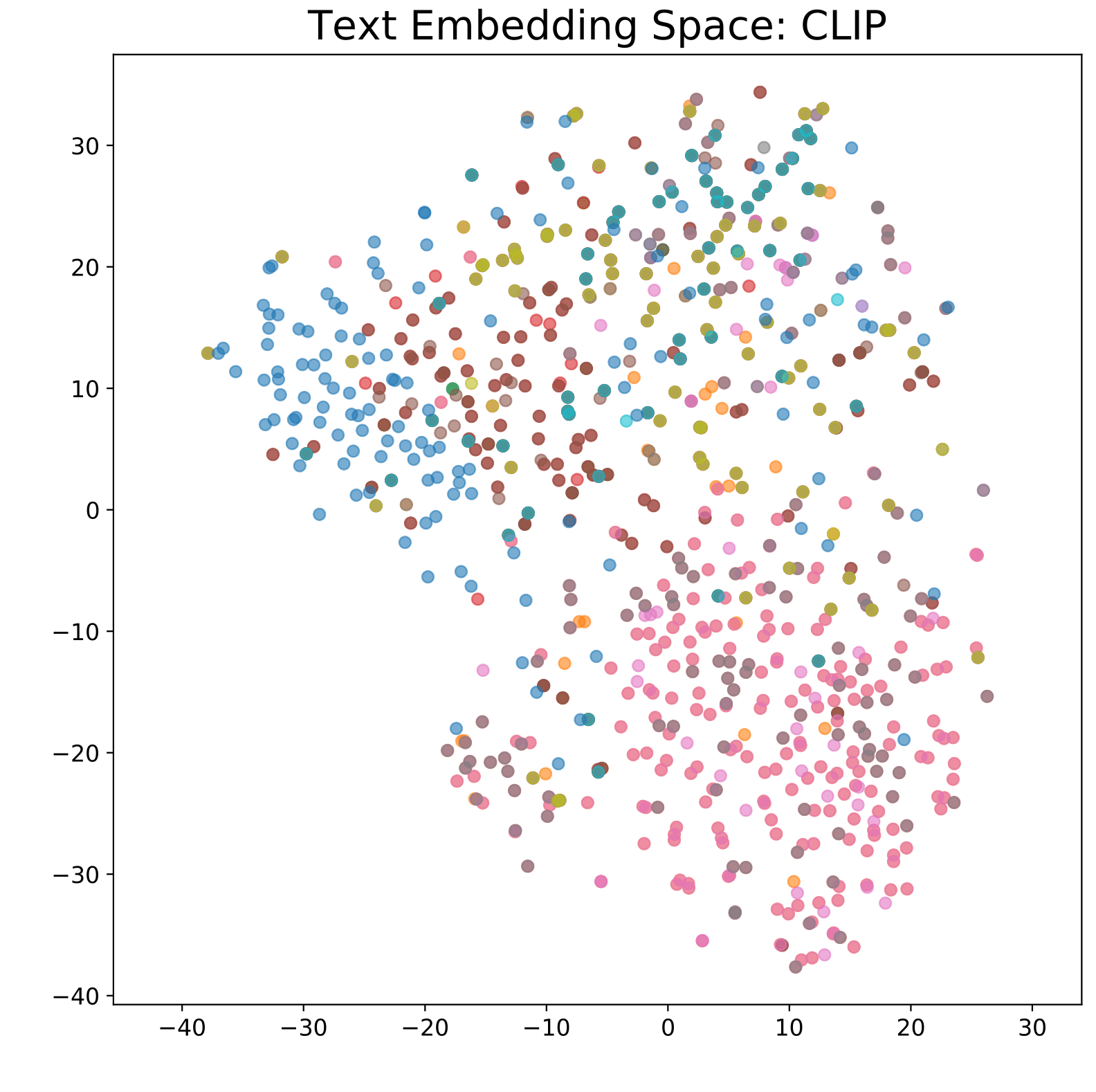}
    \label{fig:tsne_clip_text}
\end{subfigure}
\hfill
\begin{subfigure}[t]{0.247\textwidth}
    \centering
    \includegraphics[width=\textwidth]{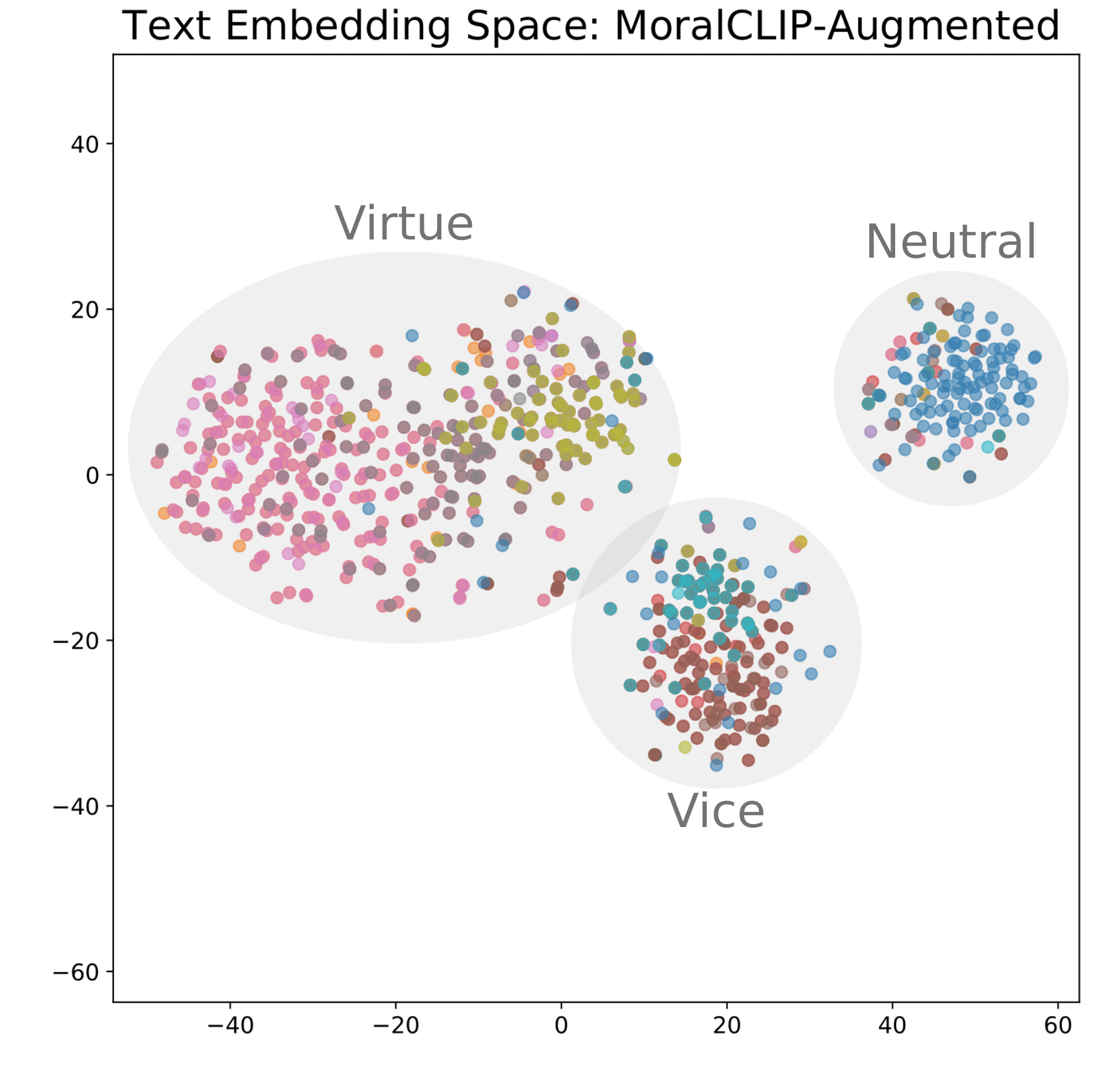}
    \label{fig:tsne_moralclip_text}
\end{subfigure}

\begin{center}
\begin{subfigure}[b]{0.7\textwidth}
    \centering
    \includegraphics[width=\textwidth]{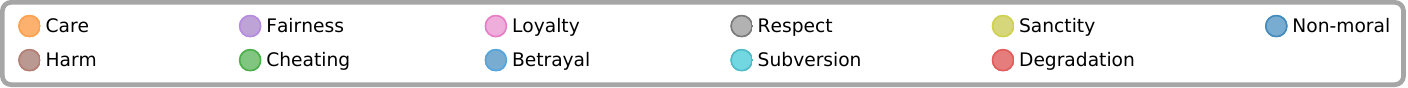}
    \label{fig:tsne_legend}
\end{subfigure}
\end{center}
\caption{t-SNE visualization of embedding spaces across different models and modalities. Points are colored by moral categories. Note that the moral annotations are multi-label, meaning individual samples can exhibit multiple dimensions simultaneously. Across both image and text embedding spaces, MoralCLIP demonstrates clearer separation between moral categories and better clustering of morally similar content compared to the baseline CLIP model.}

\label{fig:tsne_analysis}
\end{figure*}

\subsection{Methodology and Metrics}\label{subsec:metrics}

To evaluate the performance of MoralCLIP and other baselines, we developed a multi-criteria evaluation framework that assesses performance across modalities with three distinct metrics:

\noindent \textbf{Mean Average Precision (MAP).} \hspace{2mm}  MAP~\cite{manning2008ir} evaluates retrieval performance by measuring the model's capability to rank morally similar content based on similarity scores in the embedding space. For each query item, we compute cosine similarities with all other items and rank them in descending order. We consider retrieved items relevant if they share at least one moral label with the query. MAP is particularly suitable for our setting, as it rewards models that retrieve morally aligned content in earlier positions of the similarity-based ranking.

\noindent \textbf{Discriminative Power (DP).} \hspace{2mm} DP quantifies moral category separation by computing the ratio of intra-class similarity (items which share at least one moral label) to inter-class similarity (items with no shared labels). Higher values indicate stronger within-class coherence and greater between-class separation~\cite{wu2017sampling}.

\noindent \textbf{Silhouette Score.} \hspace{2mm} Silhouette score~\cite{rousseeuw1987silhouettes} assesses whether embeddings are grouped by moral polarity. We simplify our multi-label space to three broader categories (\textit{virtue}, \textit{vice}, and \textit{neutral}) to test whether the embedding space reflects these core moral distinctions. 
 
 To ensure statistical reliability, we report standard errors for all metrics computed using bootstrap resampling with 1000 iterations.

\section{Results and Discussion} 
In this section, we start by discussing MoralCLIP's quantitative results and then proceed to examine the quality of the moral data augmentation process. We conclude with a qualitative analysis.

\subsection{MoralCLIP}
We first optimized the moral loss weight ($\lambda$) for models with \textit{explicit} supervision by varying $\lambda$ from 0.1 to 0.5 with a step of 0.1. 
We then selected the best epoch-$\lambda$ combination based on validation performance. 
Values of $\{0.4,0.5\}$ consistently yielded optimal results with minimal variation between them.
Explicit moral training improved over the CLIP baseline irrespective of $\lambda$, indicating the approach is robust to moderate variations in the moral loss weight. 

Table~\ref{tab:moral_clip_metrics} presents a comprehensive evaluation of MoralCLIP variants across multiple metrics and modalities. 
These results demonstrate that \textbf{CLIP + Moral Images} (the \textit{implicit} setting, Section~\ref{subsec:implicit}) outperforms both CLIP and Safe-CLIP, underscoring the importance of simple \textit{implicit} moral supervision. This suggests that morally diverse training naturally induces latent moral structure, with stronger effects when we mix captions and images from the same category (MFT Swapper). Notably, Safe-CLIP---a fine-tuned CLIP-Large-14 model---fails to achieve meaningful moral organization despite its larger architecture. This may suggest that Safe-CLIP's methodology designed to map inappropriate content to safer embedding regions inadvertently diminishes the moral distinctions that our approach seeks to preserve and enhance.

\textbf{MoralCLIP Augmented}, with \textit{explicit} moral alignment, consistently outperforms \textit{implicit} approaches and surpasses the safety-focused Safe-CLIP baseline, achieving the best overall performance. This variant shows remarkable gains across all metrics, with MAP scores increasing by 29.62 percentage points for images (from 42.06\% to 71.68\%) and 23.04 for text (from 38.73\% to 61.77\%) when compared to the pretrained version. This improvement indicates that integrating moral similarity directly into the contrastive learning objective effectively reorganizes the embedding space along moral dimensions. This reorganization is visually confirmed through our analysis of the embedding space, which shows that MoralCLIP creates more coherent moral clusters (Figure~\ref{fig:tsne_analysis}). While silhouette scores remain modest---reflecting the context-dependent nature of moral content---the substantial improvements in MAP and DP metrics demonstrate that MoralCLIP effectively captures moral similarity relationships.

\begin{figure*}[t]
    \centering
    \includegraphics[width=.9\linewidth]{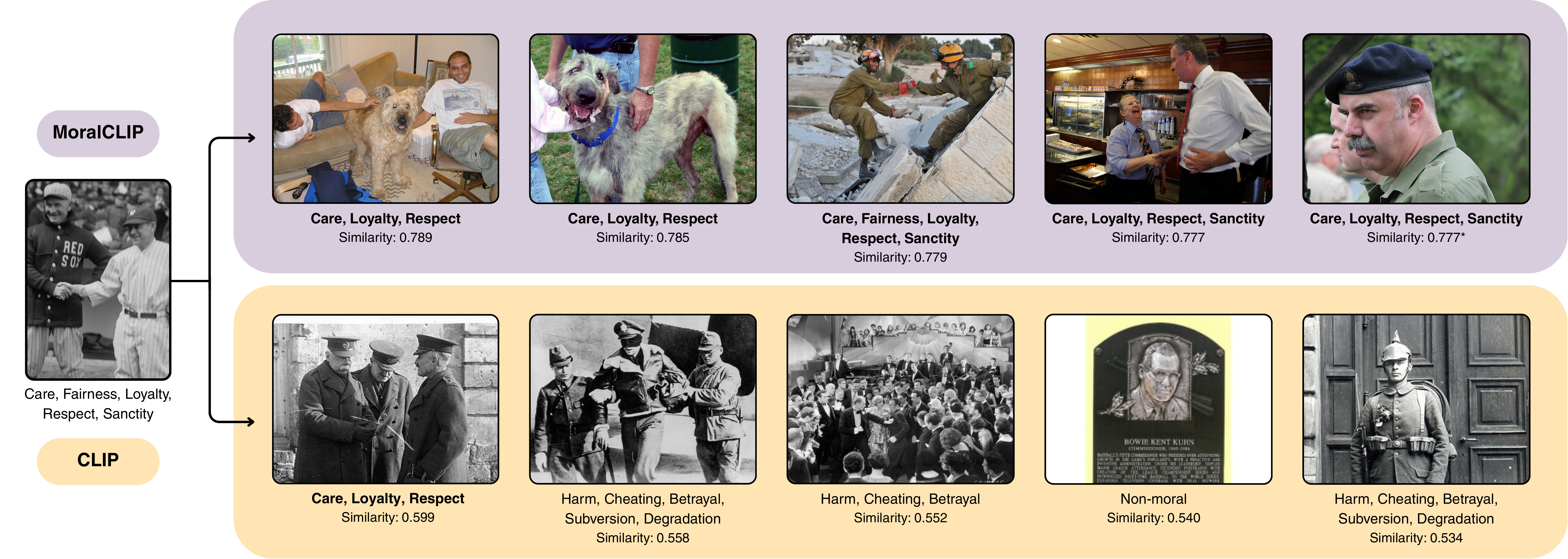}
    \caption{Image-to-Image retrieval comparison between MoralCLIP and CLIP models on the test set. MoralCLIP retrieves similar images depicting human connection across diverse contexts, while CLIP focuses on low-level visual features like color scheme and formal posing. Similarity scores represent cosine similarity. The moral labels in bold match the query's label. Throughout our figures, values marked with * indicate rounding approximations where actual values differ slightly.}
    \label{fig:retrieval_I2I}
\end{figure*}

A closer comparison across model types reveals that the optimal strategy differs between training paradigms. In the \textit{implicit} setting (CLIP + Moral Images), MFT Swapper variants outperform all baselines, suggesting that mixing content within moral categories helps the model discover latent moral structure. However, with \textit{explicit} supervision (MoralCLIP), standard augmentation surpasses MFT Swapper. This indicates that when moral labels are provided explicitly, maintaining semantic coherence matters more than moral diversity. This finding aligns with~\citet{park_2024}, who show that moral pluralism is challenging to deduce via self-supervision alone and typically requires explicit labels. Our results similarly suggest that \textit{explicit} moral supervision enables clearer distinction between moral dimensions.
Across all models and metrics, the image modality consistently outperforms text. This pattern aligns with the well-documented "modality gap", a spatial separation between image and text embeddings that emerges inherently from contrastive training~\cite{mindthegap, mitigatethegap}, suggesting that our moral alignment training, while beneficial overall, may have amplified this modality imbalance. This may also stem from visual content often containing more explicit moral cues than text~\cite{picture_this, smid, maffs, socialmoral} and limitations of CLIP's text encoder, which include a short effective input length~\cite{Long-CLIP} and limited capacity for processing subtleties in language~\cite{sitingli_2025}. This asymmetry suggests that visual content may serve as a stronger signal for moral alignment, while simultaneously highlighting the need for improved text encoders capable of handling longer, subtler moral narratives. Despite this modality imbalance, our cross-modal results demonstrate that MoralCLIP achieves robust bidirectional alignment---an essential property for consistent moral interpretation across input modalities.

Altogether, our results demonstrate that MoralCLIP successfully captures moral dimensions in multimodal representations, with \textit{explicit} supervision and appropriate augmentation strategies showing the greatest promise for developing systems that can reliably encode moral content across modalities.

\subsection{Morally-Grounded V\&L Data Augmentation}\label{subsec:vmc_agree}
To augment the dataset of image-text pairs with moral labels, we leveraged the \emph{Visual Moral Compass} classifier (Section~\ref{sec:visual_moral_compass}), which demonstrates strong and consistent performance across various moral foundations, enabling scalable and effective automated labeling for MoralCLIP training.

To assess the reliability of these automated annotations, we conducted a human annotation study using a subset of 200 images from our dataset. Twelve annotators participated across four batches of 50 images each, with three annotators per batch. One annotator was excluded due to insufficient response variability ($\sigma$=0.179). We report inter-annotator agreement using Krippendorff's $\alpha$, and human-classifier agreement using Cohen's $\kappa$ against majority labels (Table~\ref{tab:agreement_results}).

The results reveal substantial variation in human agreement across moral foundations ($\alpha=0.184-0.417$), with \textit{Care} showing the highest consensus and \textit{Fairness} the lowest---a pattern that closely mirrors our classifier's performance hierarchy, where \textit{Care} achieved the strongest F1 ($0.84$) and \textit{Fairness} the weakest ($0.71$), with other foundations clustering around $0.81$. These trends are consistent with findings in moral psychology that suggest certain foundations are less culturally variable than others~\cite{MFT, Graham2011}. For instance, \textit{Care} is often regarded as a universal moral concern, whereas \textit{Fairness} and \textit{Authority} show greater cultural differences and ideological divergence~\cite{atari_weird_2023, MFT, Graham2011}. Moreover, the agreement levels we observe are consistent with---and in some cases exceed---those reported in prior moral annotation efforts, including MFTC~\cite{MFT_twitter} and MFRC~\cite{MFT_Reddit}. 
\begin{table}[t]
    \centering
    \caption{Inter-annotator agreement (Krippendorff’s $\alpha$) and human-classifier agreement against majority vote (Cohen's $\kappa_{\text{maj}}$). Consensus Coverage (CC) shows the percentage of examples with annotator consensus.}
    \label{tab:agreement_results}
    \begin{tabular}{@{}lccc@{}}
        \toprule
        \textbf{Foundation} & \textbf{IAA ($\alpha$)} & \textbf{Model ($\kappa_{\text{maj}}$)} & \textbf{CC} \\
        \midrule
        Care     & 0.417 $\pm$ 0.029 & 0.451 $\pm$ 0.180 & 85.1\% \\
        Fairness    & 0.184 $\pm$ 0.145 & 0.233 $\pm$ 0.083 & 94.6\% \\
        In-group  & 0.293 $\pm$ 0.041 & 0.338 $\pm$ 0.171 & 88.4\% \\
        Authority  & 0.217 $\pm$ 0.137 & 0.024 $\pm$ 0.091 & 83.7\% \\
        Purity & 0.397 $\pm$ 0.190 & 0.216 $\pm$ 0.126 & 93.8\% \\
        \bottomrule
    \end{tabular}
\end{table}
More importantly, model-human agreement followed a similar pattern, with highest alignment on \textit{Care} ($\kappa_{\text{maj}} = 0.451$) and lowest on \textit{Authority} ($\kappa_{\text{maj}} = 0.024$), reflecting the differing levels of annotator consensus. These results, combined with high consensus coverage (83.7\%–94.6\%), validate that our classifier captures genuine moral patterns suitable for MoralCLIP training. Detailed annotation workflow and instructions are available in Appendix \ref{app:human_annotations}.

\begin{figure*}[t]
    \centering
    \includegraphics[width=1.0\linewidth]{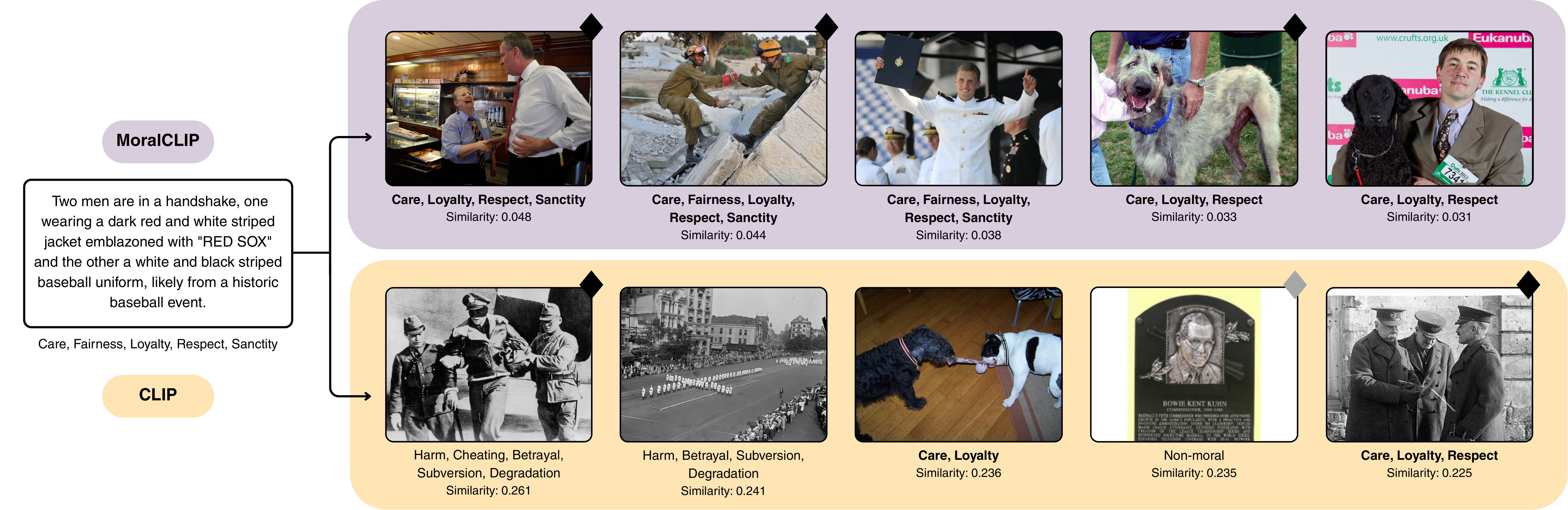}
    \caption{Text-to-Image retrieval comparison between MoralCLIP and CLIP models on the test set. Given a query of a handshake scene, MoralCLIP retrieves images depicting moral themes of care and respect, while CLIP mostly retrieves achromatic images with historical elements. \ding{117} indicates images also retrieved in our image-to-image evaluation, while {\color{gray} \ding{117}} indicates same-position retrievals. Similarity scores represent cosine similarity. The moral labels in bold match the query's label.}
    \label{fig:example_T2I}
\end{figure*}

\subsection{Qualitative Analysis}
\subsubsection{Embedding Space Visualization}\label{subsubsec:tsnes}

Figure~\ref{fig:tsne_analysis} depicts CLIP and MoralCLIP embedding spaces across both image and text modalities. Standard CLIP (first and third panels) produces scattered embeddings with limited moral clustering, reflecting its purely semantic training objective. Virtue, vice, and neutral examples are mixed throughout the space, with no discernible moral clustering structure. Incidentally, there is limited grouping of certain moral categories, likely due to semantically related concepts, such as weapons (related to \textit{Harm}) or animals (relating to \textit{Care}) already clustering in CLIP's embedding space. 

In contrast, MoralCLIP-Augmented (second and fourth panels) displays improved moral organization with clear virtue-vice separation across both modalities. Although consistent, this moral clustering effect is more pronounced in the image embedding space than in text, aligning with our quantitative findings (Table~\ref{tab:moral_clip_metrics}). This visualization indicates that moral supervision successfully transforms semantically-organized representations into ones that are organized along moral dimensions.

\subsubsection{Retrieval Analysis}
To further investigate how MoralCLIP's embedding space differs from standard CLIP, we analyze retrieval performance using MoralCLIP-Augmented. Using a consistent query across all four modality combinations, Figure~\ref{fig:retrieval_I2I} depicts the image-to-image retrieval comparison and Figure~\ref{fig:example_T2I} showcases the text-to-image retrieval example. Retrieval is performed using cosine similarity in the learned embedding space without any explicit use of moral labels during the retrieval process. Additional cross-modal retrieval examples are included in Appendix~\ref{app:retrieval_results}.

The results reveal a fundamental difference in how the two models interpret content across modalities. We illustrate this with a query depicting a handshake, symbolizing cooperation and respect. 
In image-to-image retrieval, MoralCLIP retrieves diverse yet morally aligned content: dog companionship (nurturance), collaborative labor (cooperation), two individuals in conversation (connection), and a contemplative military serviceman (respect). Similarly, in text-to-image retrieval (Figure~\ref{fig:example_T2I}), MoralCLIP retrieves images emphasizing themes of care and human connection rather than literal visual matches. This demonstrates that MoralCLIP's similarity metric is driven by moral associations rather than surface-level patterns, with all retrieved content showing significant moral label overlap with the query. 
In contrast, CLIP's results clearly reflect its reliance on surface-level characteristics across both tasks. In image retrieval, all results are black-and-white, suggesting the model prioritizes the monochromatic nature of the query over its semantic content. In text-to-image retrieval, CLIP similarly retrieves black-and-white historical imagery, responding to the "historic" descriptor mentioned in the query text. This aligns with prior findings that CLIP associates grayscale imagery with historical contexts~\cite{Offert2023, blind_dates2023} and struggles with fine-grained detail~\cite{krojer_image_2022, clip_finegrained}. While CLIP's literal interpretation is technically accurate, it focuses on descriptive and visual characteristics rather than the underlying moral significance of human interaction, missing the deeper themes of cooperation and respect that transcend historical context.

\section{Conclusions and Opportunities}

As AI systems permeate everyday life, the ability to understand moral dimensions becomes essential. 
In this paper, we introduced the first framework for multimodal moral interpretation, advancing our understanding of ethical content across visual and textual media. The key contributions and takeaway lessons are as follows:

\noindent \textbf{MoralCLIP.} \hspace{2mm} The MoralCLIP embedding space moves us toward AI systems capable of recognizing, and eventually reasoning about, moral dimensions grounded in Moral Foundations Theory. Our results reveal that \textit{explicit} moral supervision outperforms \textit{implicit} approaches, strongly indicating that moral understanding requires explicit guidance rather than emerging naturally from general vision-language training. 
The framework enables bidirectional cross-modal moral understanding, highlighting opportunities for richer moral text analysis beyond simple descriptive captions.

\noindent \textbf{V\&L alignment with MFT.} \hspace{2mm} Overall, our results demonstrate that our moral training approach aligns representations with morally-relevant dimensions, enabling recognition of moral content rather than superficial visual traits. While this doesn't solve all aspects of moral understanding, it represents a meaningful shift in what the model attends to. This reorientation toward moral dimensions represents progress in developing multimodal systems capable of moral reasoning, highlighting the potential of value-aligned embedding spaces for future advances in this domain. Future work could further explore moral similarity metrics beyond the Jaccard index that better capture nuanced relationships between moral foundations that discrete set overlap measures cannot fully represent.

\noindent \textbf{Weak Labeling.} While our automated labeling approach via the \emph{Visual Moral Compass} enables scalable dataset creation, it introduces potential noise compared to expert annotation. However, our human evaluation study demonstrates that our classifier achieves reasonable agreement with human annotators across most moral foundations. Additionally, our approach leverages SMID's expert-validated annotations as the foundation for training the \emph{Visual Moral Compass}, ensuring our automated labels build upon established moral ground truth. Even so, future work would benefit from direct human moral annotation of visual and textual content.

\noindent \textbf{Cultural and Demographic Bias.} \hspace{2mm} While MFT posits that its five moral foundations are universal across cultures, it also acknowledges that different groups prioritize them differently. Despite SMID's use of thousands of annotators to ensure demographic diversity, the final dataset aggregates these annotations into single moral labels, effectively averaging out cultural variation in moral judgment. This aggregation approach enables stable training labels but erases variation in moral interpretation. 
This bottleneck appears particularly complex to handle within current annotation frameworks, as preserving cultural diversity would require maintaining multiple, potentially conflicting labels for the same content, fundamentally changing how we approach both dataset construction and model training.

In summary, MoralCLIP bridges multimodal learning with Moral Foundation Theory, opening a new research direction at the intersection of multimedia understanding and computational ethics. By embedding moral foundations into vision and language models, our work lays the groundwork for future systems that are not only semantically rich and multimodal, but also capable of engaging with the moral dimensions of human communication.

\section*{Acknowledgments}
This work is supported by UID/04516/NOVA Laboratory for Computer Science and Informatics (NOVA LINCS) with the financial support of FCT/IP. It was also supported by the AMALIA project, funded by FCT/IP in the context of measure RE-C05-i08 of the Recovery and Resilience Program.

\bibliographystyle{ACM-Reference-Format}
\bibliography{bibliography}

\clearpage
\appendix
\section{Multimodal Moral Data Processing}
\label{app:vmc_smid_preprocess}
\begin{figure}[h]
    \centering
 \includegraphics[width=\linewidth]{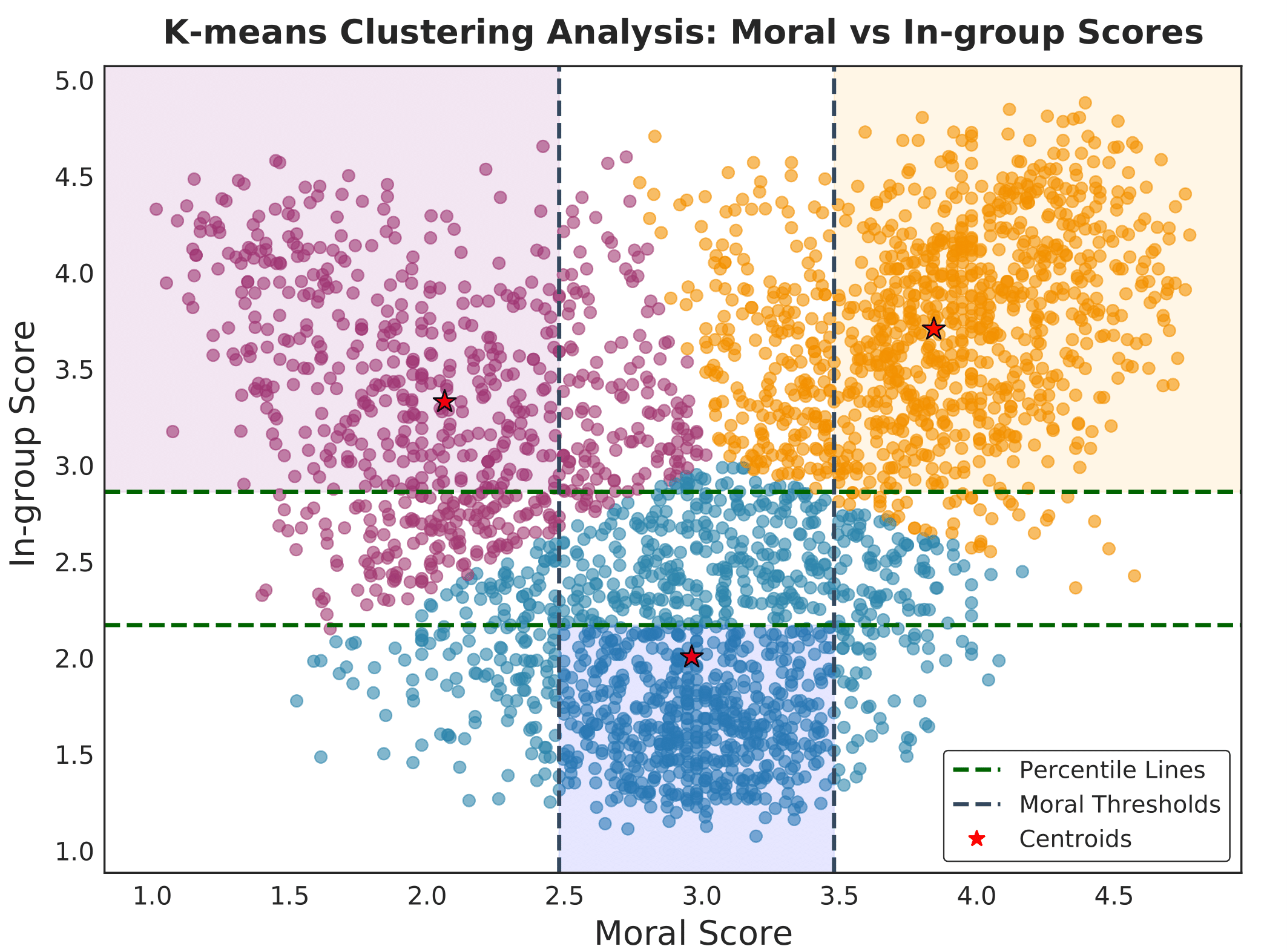}
    \caption{Classification of SMID images into moral categories based on moral valence and relevance scores. Images are categorized as \textit{Vice} (purple region, negative moral valence), \textit{Virtue} (orange region, positive moral valence), or \textit{Neither} (blue region, neutral moral valence). Thresholds are set at moral scores < 2.5 (negative), > 3.5 (positive), and relevance scores < 2.15 (low), > 2.84 (high) to exclude ambiguous boundary cases.}
    \label{fig:figure1_kmeans}
\end{figure}

We adapted SMID's annotations to classify each image into one of three categories---\textit{Virtue}, \textit{Vice} or \textit{Neither}---based on its moral ratings (Figure~\ref{fig:figure1_kmeans}). Thresholds for moral valence ($x$) were
defined by the authors (negative: $x<2.5$, positive: $x>3.5$).
Similarly, in order to exclude images that lie close to cluster boundaries, we excluded images with ambiguous relevance scores. Thresholds for low ($y < 2.15$) and high ($y > 2.84$) relevance were determined using percentile-based distributions. This approach operates under the assumption that morally positive images are associated with virtue, while morally negative images correspond to vice within a specific moral foundation. While this simplification aligns with common interpretations of MFT~\cite{MFT, MoralBERT}, we acknowledge that some images may evoke ambiguous moral responses, which we attempt to address by excluding boundary cases.
Figure~\ref{fig:figure1_kmeans} illustrates this classification process. Images in the purple region were labeled as \textit{Vice} (e.g. \textit{Betrayal }in the \textit{In-group} foundation), in the orange region as \textit{Virtue} (e.g. \textit{Loyalty}), and in the blue region as \textit{Neither}. All other foundations follow a nearly identical pattern, as reported by the SMID authors~\cite{smid}.
To determine inclusion in the final dataset, we evaluated each image across all five moral foundations. Images excluded from fewer than five foundations were retained, as their relevance and morality scores aligned with at least one moral foundation. Conversely, images consistently classified outside the defined regions across all dimensions were removed. This process refined the dataset to 2,401 images, which were used for model fine-tuning. For multi-label classification purposes, we mapped these classifications to numerical labels, to guarantee alignment with the structured Virtue-Vice dichotomy within MFT, while preserving the dataset's granularity and multi-dimensional annotations.

\section{Visual Moral Compass: Implementation Details}\label{app:vmc}

\subsection{Training and Loss Formulation}\label{app:vmc_loss}

\begin{figure}[t]
    \centering
    \includegraphics[width=\linewidth]{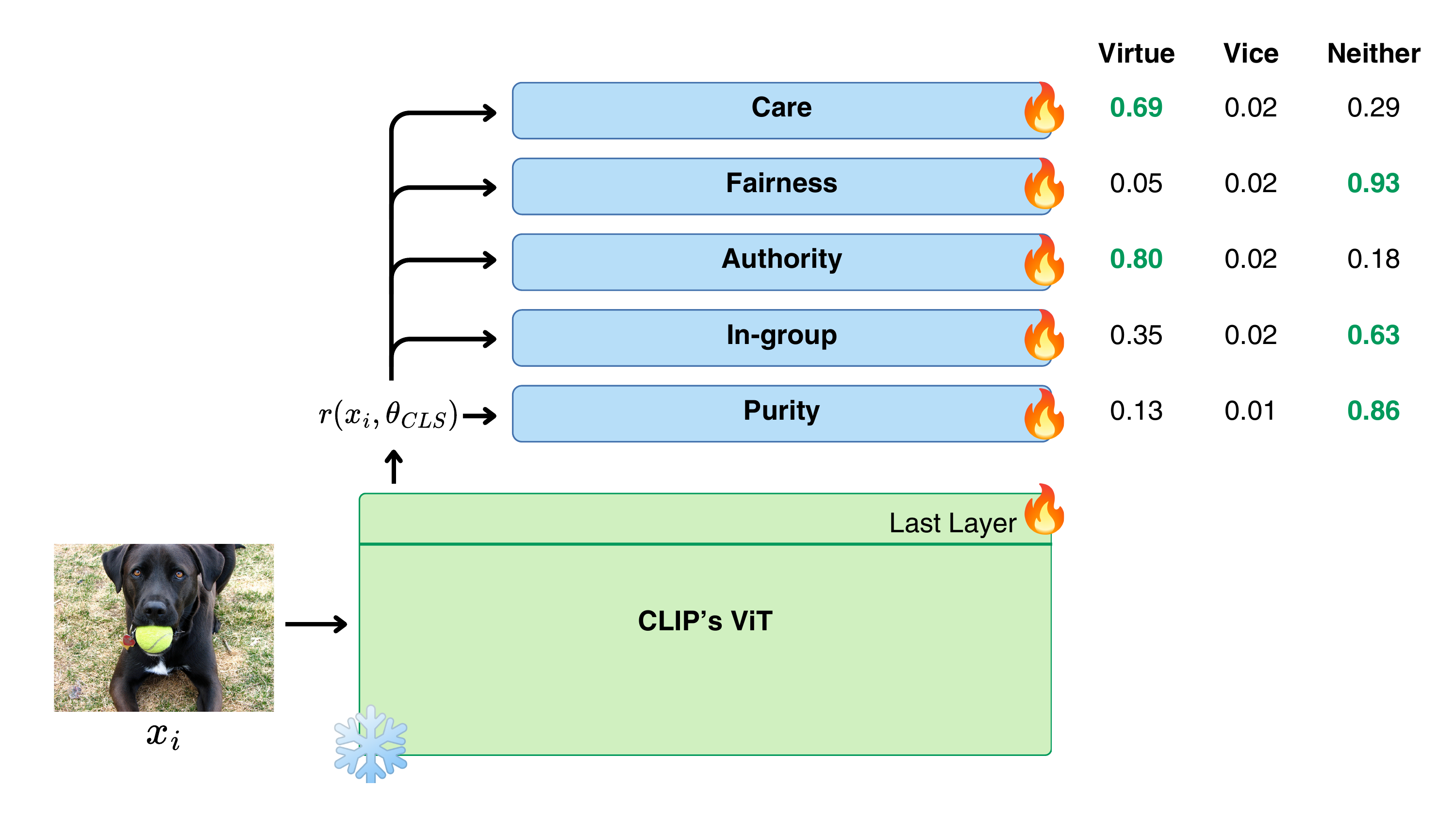}
    \caption{Overview of the \emph{Visual Moral Compass} architecture.}
    \label{fig:visual_moral_compass}
\end{figure}

We fine-tune the CLIP (ViT-B/16)~\cite{CLIP} encoder, adapting its vision encoder for a multi-label classification task aligned with the moral foundations defined by the MFT~\cite{MFT}. Given an image $x_i$, the encoder extracts a feature vector $r(x_i, \theta_{\text{CLS}})$, derived from the \textsc{[CLS]} token. This embedding is passed to five independent classifier heads $f_i$, one per moral foundation (Figure~\ref{fig:visual_moral_compass}). Each head outputs a probability distribution over three classes: \textit{Virtue}, \textit{Vice} or \textit{Neither}:
\begin{equation}
P_{\theta_{\text{CLS}}, \theta_{y_i}}(y_i | x) = softmax(f_i(r(x,\theta_{\text{CLS}}); \theta_{y_i}))
\end{equation}
where $\theta_{\text{CLS}}$ represents the trainable parameters of the final layer of the encoder, and $\theta_{y_i}$ are the parameters of classifier $f_i$. Each label $y_i \in \{0,1,2\}$ corresponds to a class within foundation $i$.

\paragraph{Optimization} The model jointly optimizes both $\theta_{\text{CLS}}$, and $\theta_{y_i}$. While the encoder provides a shared visual representation, allowing interaction between moral foundations, independent classifier heads enforce virtue-vice exclusivity within each moral axis. The total loss is computed as the sum of the cross-entropy losses across all five foundations:
\begin{equation}
\mathcal{L}(\theta_{\text{CLS}}, \theta_{y}) = \sum_{i=1}^{k} \text{CrossEntropy}(P_{\theta_{\text{CLS}}, \theta_{y_i}}(y_i | x), y_i),
\end{equation}
where $k=5$ corresponds to the five moral foundations defined by MFT~\cite{MFT}, $P_{\theta_{\text{CLS}}, \theta_{y_i}}(y_i | x)$ the predicted probability distribution over the three possible classes for the $i$-th moral foundation, and $y_i$ the true class label.
\begin{table*}[ht]
    \centering
    \caption{Summary of Test Set labels for all moral dimensions.}
    \label{tab:smid-test-labels}
    \begin{tabular}{lcccccc}
        \toprule
        \multirow{2}{*}{\textbf{Label}} & \textbf{Care} & \textbf{Fairness} & \textbf{Loyalty} & \textbf{Respect} & \textbf{Sanctity} & \multirow{2}{*}{\textbf{Neither}} \\
        & \textbf{Harm} & \textbf{Cheating} & \textbf{Betrayal} & \textbf{Subversion} & \textbf{Degradation} & \\
        \midrule
        \textbf{Count} & 81/53 & 40/29 & 86/31 & 85/30 & 56/46 & 87 \\
        \bottomrule
    \end{tabular}
\end{table*}

\subsection{Hyperparameters}\label{app:vmc_imp_details}
The \textit{Visual Moral Compass} was trained on a single A100 40GB GPU for 20 epochs using the Adam optimizer with a learning rate of $1\times10^{-4}$. To dynamically adjust the learning rate, we employed a \textit{ReduceLROnPlateau} scheduler, reducing the learning rate by a factor of 0.1 if the validation F1 score does not improve for three consecutive epochs. We apply early stopping with a patience of 8 epochs after the initial 10 epochs. We set the batch size to 32. The curated SMID dataset was used for model fine-tuning, using 10\% of the data for validation and other 10\% for testing. We showcase the test set labels from SMID in Table~\ref{tab:smid-test-labels}.

\section{Visual Moral Compass: Performance}\label{app:vmc_performance}
To evaluate the performance of the \textit{Visual Moral Compass} across the 10 moral dimensions, we report macro-averaged Precision, Recall, and F1. 
As shown in Table~\ref{tab:test_bertdst}, the \textit{Visual Moral Compass} achieves strong overall performance, with an average F1 of 0.796 across all moral foundations, indicating reliable single-dimension prediction. The \textit{Care} foundation performs best, likely due to the intuitive recognition of pro-social behaviors~\cite{MFT, narrative}. Conversely, \textit{Fairness} shows weaker results, which is consistent with the challenge of representing abstract moral concepts visually~\cite{weber_2018, eMFD}

\begin{table}[h]
\centering
\caption{Performance metrics for the classifier heads. Metrics are computed using macro-averaging across the three classes (\textit{Virtue}, \textit{Vice}, \textit{Neither}) within each moral foundation.}
\label{tab:test_bertdst}
\resizebox{\columnwidth}{!}{%
\begin{tabular}{lcccc}
\toprule
\multirow{2}{*}{\textbf{Classifier}} & \multirow{2}{*}{\textbf{Accuracy}} & \multicolumn{3}{c}{\textbf{Metrics}} \\
\cmidrule(lr){3-5}
& & \textbf{Precision} & \textbf{Recall} & \textbf{F1-Score} \\
\midrule
Care       & 0.851 & 0.853 & 0.838 & 0.844 \\
Fairness   & 0.805 & 0.723 & 0.705 & 0.712 \\
In-group   & 0.838 & 0.820 & 0.797 & 0.807 \\
Authority  & 0.847 & 0.831 & 0.796 & 0.811 \\
Purity     & 0.830 & 0.818 & 0.797 & 0.807 \\
\cmidrule(lr){1-5}
\textbf{Average}    & \textbf{0.834} & \textbf{0.809} & \textbf{0.786} & \textbf{0.796} \\
\bottomrule
\end{tabular}%
}
\end{table}

While recent work acknowledges the pluralistic nature of moral reasoning~\cite{park_2024, MoralBERT, mformer}, most approaches still rely on binary classifiers that treat each foundation independently, limiting their ability to capture inter-foundation relationships. Our own model is similarly constrained: due to limited training data, we fine-tuned only the final CLIP encoder layer, which may hinder its capacity to model foundation interdependencies. Nevertheless, the \textit{Visual Moral Compass} performs robustly in classifying individual foundations and provides reliable annotations for building our multimodal dataset and embedding space. We anticipate that larger-scale moral datasets will enable full encoder training and more effectively capture foundation interdependencies, ultimately improving joint moral inference.

\section{MoralCLIP Training Details}\label{app:training_parameters}
Table~\ref{tab:train_config} summarizes the training parameters, data augmentation strategies, and MFT mixing configurations used across all model variants. All experiments were conducted on a single A100 40GB GPU, using identical optimization settings to ensure fair comparison. 

\begin{table}[ht]
\centering
\caption{Training Configuration Parameters}
\label{tab:training-config}
\begin{tabular}{@{}ll@{}}
\toprule
\textbf{Parameter} & \textbf{Value} \\
\midrule
Base Model & \texttt{clip-vit-base-patch16} \\
Training Epochs & 10 \\
Batch Size & 64 \\
Learning Rate & 1e-5 \\
Weight Decay & 0.01 \\
Optimizer & AdamW \\
LR Scheduler & Cosine Annealing \\
Temperature ($\tau$) & 0.07 \\
Evaluation Split & 5\% \\
\midrule
\multicolumn{2}{@{}l@{}}{\textbf{Augmentation Parameters}} \\
\midrule
Augmentations per Sample & 4 \\
Rotation Range & $\pm$15° \\
Brightness/Contrast/Color & $\pm$20\% \\
Gaussian Blur Radius & 0.5--1.5 \\
\midrule
\multicolumn{2}{@{}l@{}}{\textbf{MFT Swapper Parameters}} \\
\midrule
Mix Percentage & 75\% \\
Mix Types & Image or Text \\
Selection Strategy & Random within moral groups \\
Max Swaps (\textbf{Mild}) & 500 per moral group \\
Max Swaps (\textbf{Max}) & No limit \\
\bottomrule
\end{tabular}\label{tab:train_config}
\end{table}

\section{Human Annotations}\label{app:human_annotations}
\begin{figure*}[ht]
    \centering
    \includegraphics[width=0.8\linewidth]{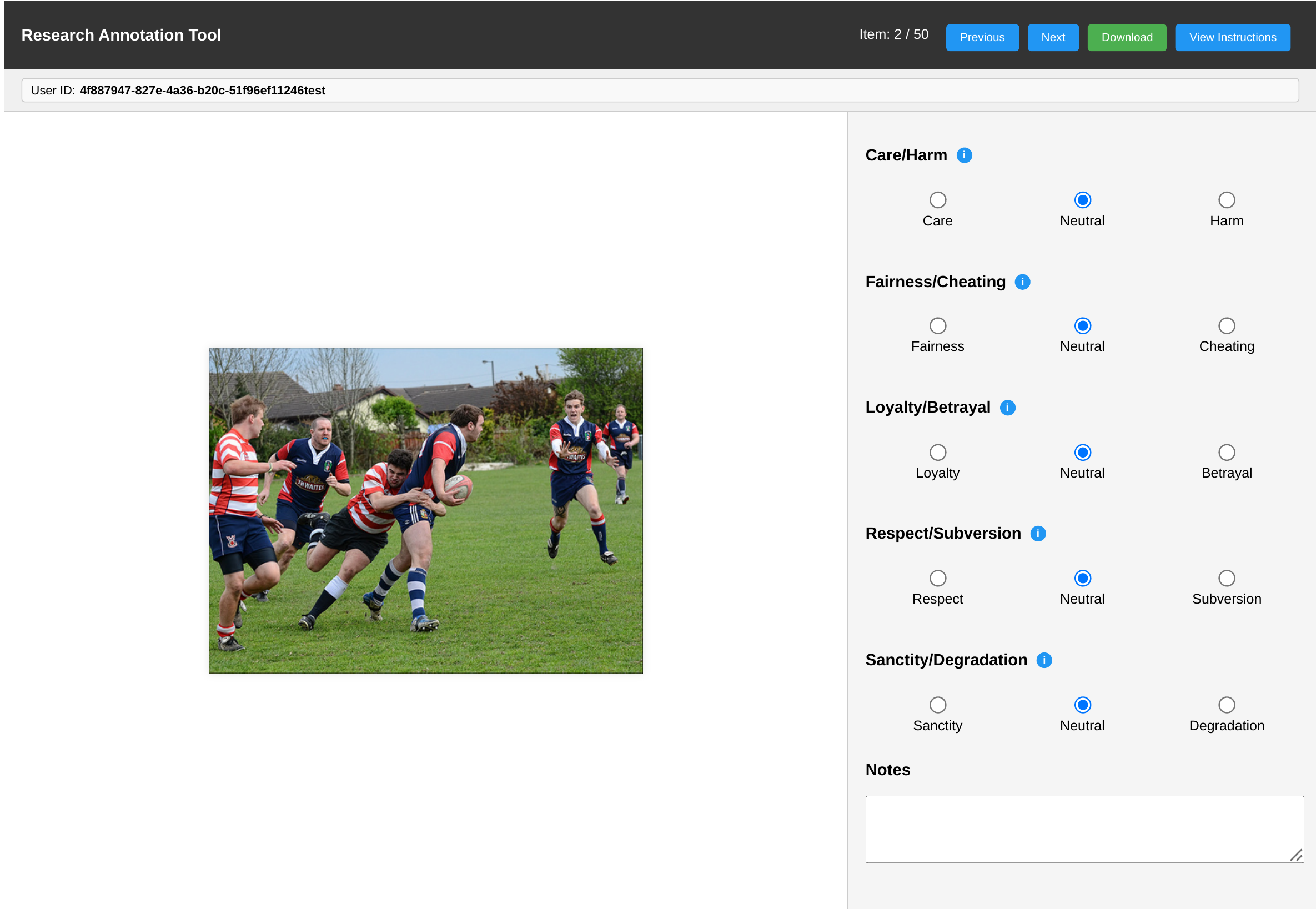}
    \caption{Human Annotation Interface: Moral Image Classification.}
    \label{fig:annotation_tool}
\end{figure*}

To collect moral foundation annotations, we developed a custom web-based annotation tool tailored to the principles of MFT. The platform allows participants to annotate images across five moral foundations: \textit{Care/Harm},\textit{ Fairness/Cheating}, \textit{Loyalty/Betrayal}, \textit{Authority/Subversion}, and \textit{Sanctity/Degradation}. Each image is presented alongside a consistent rating interface where annotators select one of three options---virtue, neutral, or vice---for each moral foundation. The platform also includes a free-text 'Notes' section where annotators could optionally clarify their interpretation of a scene or flag confusing content or technical issues. An overview of the annotation interface is shown in Figure~\ref{fig:annotation_tool}.

\begin{figure*}[ht]
    \centering
    \includegraphics[width=0.6\linewidth]{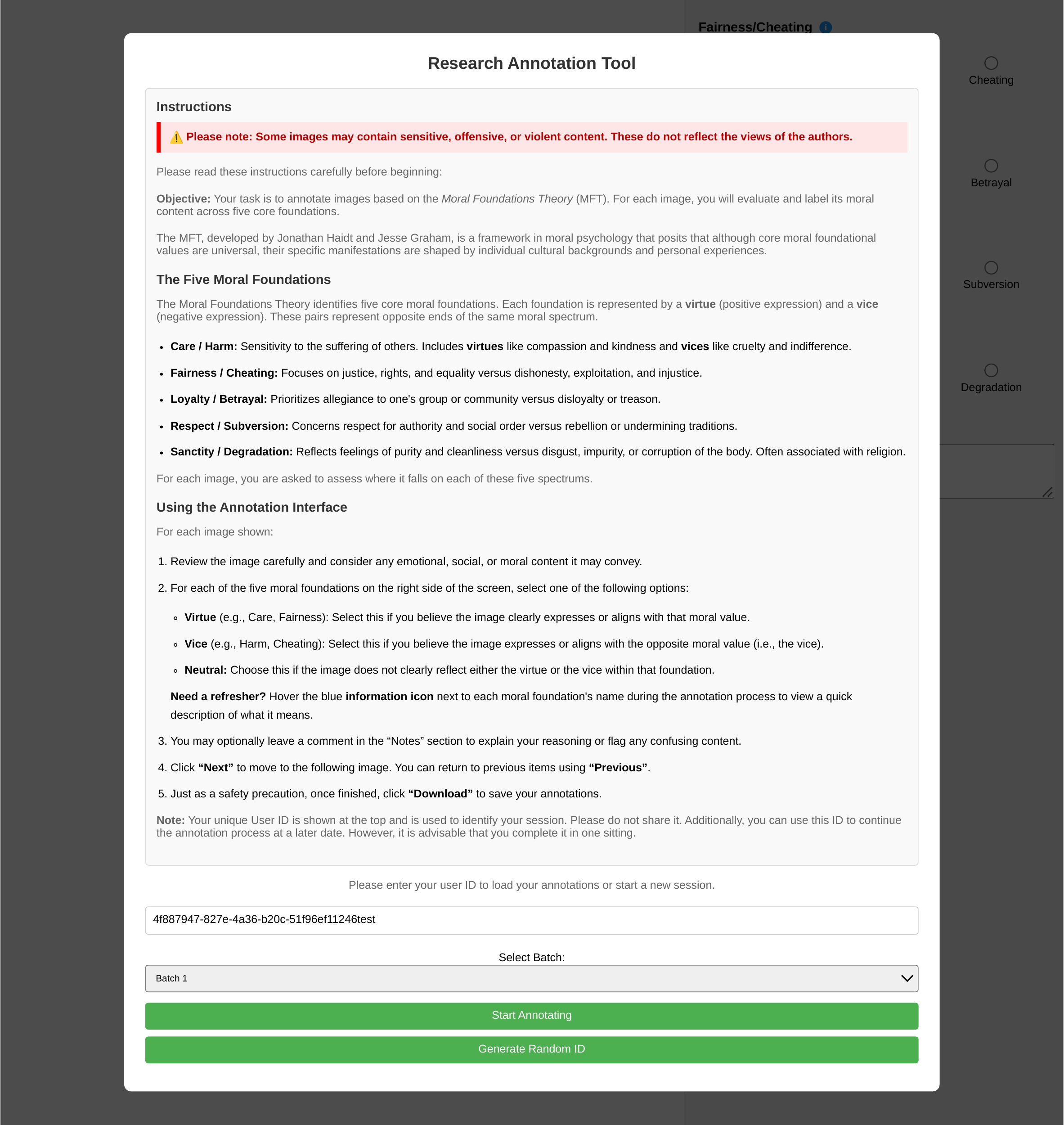}
    \caption{Initial Instruction view of the Annotation Tool.}
    \label{fig:annotation_instructions}
\end{figure*}

\paragraph{Instructions.} Annotators were first presented with detailed, structured instructions outlining the goals of the task, definitions of the five moral foundations, and step-by-step instructions on to interact with the annotation interface (Figure~\ref{fig:annotation_instructions}). The full instructions remained accessible throughout the task via a 'View Instructions' button, allowing annotators to revisit definitions or mitigate uncertainties at any point during the process. To reduce ambiguity, each foundation included an accompanying tooltip during annotation for quick reference.

To encourage fast, intuitive responses---consistent with MFT's assumption that moral judgments are often automatic and emotionally driven~\cite{MFT}---annotators were explicitly instructed to answer quickly, relying on their immediate impressions rather than deliberate reasoning.

\paragraph{Annotator Recruitment and Consent.} All annotators were university students ranging from undergraduate to PhD levels across various academic disciplines and nationalities. Annotators were aware they were contributing to an academic research project to evaluate the alignment of an image moral classifier with human moral judgment. Prior to the annotation, they received clear information about the task, the research goals and how their responses would be used. 

\section{Complete Cross-modal Retrieval Examples}\label{app:retrieval_results}
\begin{figure*}[htbp]
    \centering
    \includegraphics[width=1.0\linewidth]{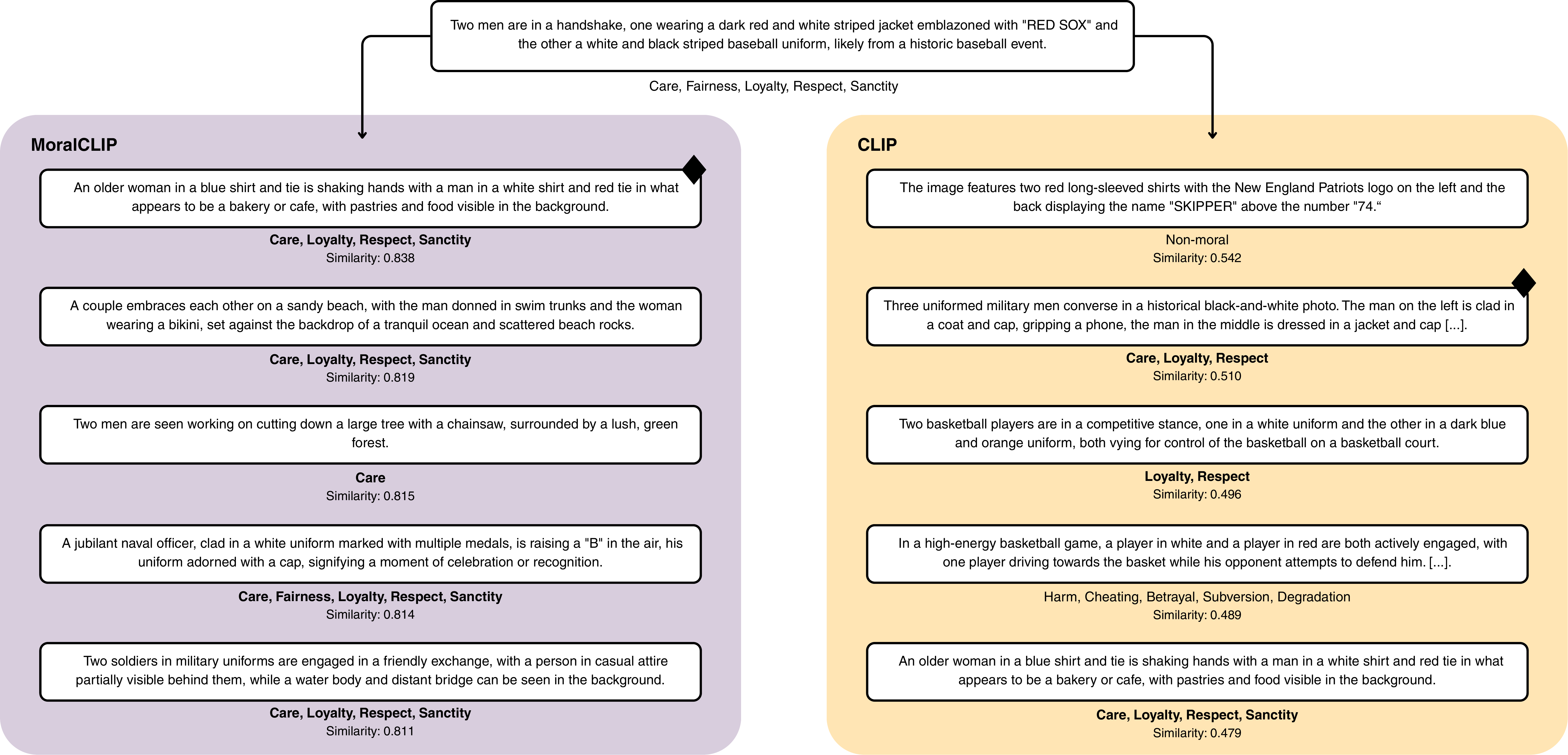}
    \caption{Text-to-Text retrieval comparison between MoralCLIP and CLIP models on the test set. Given the same handshake query, MoralCLIP retrieves morally similar descriptions across varied contexts, while CLIP retrieves text with literal or sports-related similarities. \ding{117} indicates text descriptions that correspond to images also retrieved in our image-to-image evaluation. Similarity scores represent cosine similarity. The moral labels in bold match the query's label.}
    \label{fig:example_T2T}
\end{figure*}

\begin{figure*}[htbp]
    \centering
    \includegraphics[width=1.0\linewidth]{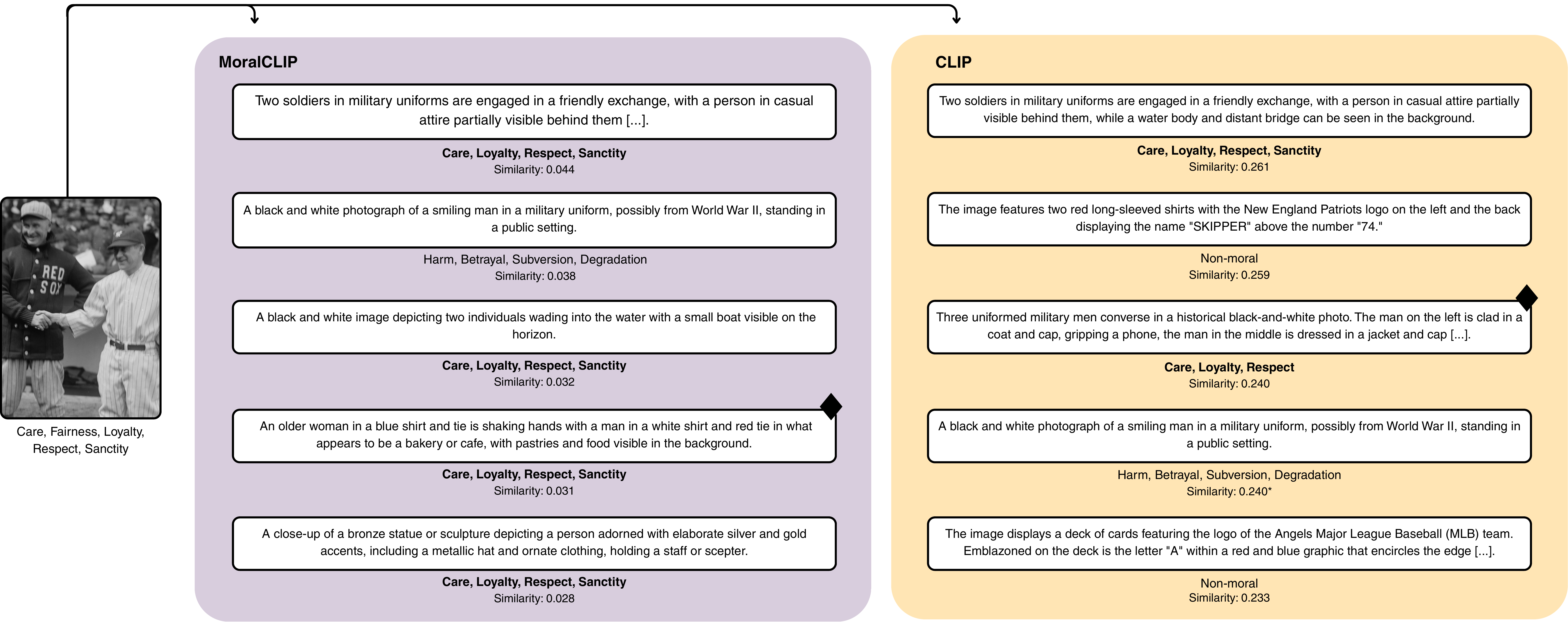}
    \caption{Image-to-Text retrieval comparison between MoralCLIP and CLIP models on the test set. Given the same query image of a handshake, MoralCLIP retrieves text descriptions emphasizing connection and respect, while CLIP retrieves descriptions focusing on visual elements like clothing and objects. \ding{117} indicates text descriptions that correspond to images also retrieved in our image-to-image evaluation. Similarity scores represent cosine similarity. The moral labels in bold match the query's label.}
    \label{fig:example_I2T}
\end{figure*}

As shown in Figures~\ref{fig:example_T2T} and \ref{fig:example_I2T}, MoralCLIP consistently demonstrates its moral alignment, retrieving data with overlapping moral labels to the query, especially when compared to standard CLIP. Interestingly, both models exhibit cross-modal consistency, but in fundamentally different ways. MoralCLIP maintains a constant "moral focus": the scene with the elderly man and woman conversing appears across tasks, accompanied by entries with positive moral labels akin to the query's. CLIP also displays consistency, often retrieving sports and military-themed content across modalities. However, the key difference lies in the organizing principle: MoralCLIP’s consistency is driven by shared moral dimensions (e.g., \textit{Care}, \textit{Respect}, \textit{Loyalty}) that transcend specific visual or textual features. Conversely, CLIP’s consistency is tied to recognizable domain categories and surface-level traits---monochromatic aesthetic and formal poses for the image modality (Figure~\ref{fig:retrieval_I2I}), sports terminology and team names for the text modality (Figure~\ref{fig:example_T2T}), and combinations of both in cross-modal settings (Figures~\ref{fig:example_I2T} and \ref{fig:example_T2I}). Thus, although both models appear to to be sensitive to modality-specific features---MoralCLIP \textit{is} a CLIP-based model, after all---MoralCLIP's alignment enables a more abstract, morally grounded connection across semantically diverse content.

In text-to-text retrieval (Figure~\ref{fig:example_T2T}), MoralCLIP's ability to identify semantically diverse scenarios that are unified by consistent moral themes is again on display, with notably high similarity scores that indicate strong moral alignment despite drastic semantic differences. This broader diversity in text-based retrievals may partly reflect the nature of the proxy labeling process. Moral labels were assigned to the original images using the \emph{Visual Moral Compass} classifier, and the generated captions inherited those labels. As a result, the moral labels in text-based evaluations reflect interpretations of visual content as captured in language, rather than direct moral assessment of the text itself. While this abstraction enables more flexible moral generalization, it may also introduce mismatches or over-generalizations, as the moral meaning of the textual descriptions may diverge from the original visual moral context. Future work could explore the integration of direct moral evaluation of textual content, or hybrid approaches that jointly model visual and textual moral cues. Furthermore, incorporating human-annotated moral labels for both modalities could help validate and refine the labeling approach, improving both reliability and interpretability.

Regarding image-to-text retrieval, both models exhibit consistent patterns, but with important differences in how they align information across modalities. Image-to-text retrieval tends to yield less coherent results than visual tasks, likely because it must map into a text space with distinct distributional properties. For CLIP, this manifests as descriptions focusing on visual elements like clothing and objects, while MoralCLIP retrieves text descriptions emphasizing connection and respect (Figure~\ref{fig:example_I2T}).

A notable difference between the models emerges in their similarity score ranges across cross-modal tasks. CLIP consistently produces higher similarity scores (0.2XX-0.5XX range), while MoralCLIP's cross-modal similarities are substantially lower (in the 0.0XX range). This disparity may result from two interacting factors: (1) our moral specialization approach that prioritizes abstract ethical understanding over literal visual-textural correspondence, and (2) the proxy labeling methodology where text encoders learn from moral labels applied to generated captions rather than direct moral annotations. This creates a representational gap where visual moral features are learned directly from images, while textual moral features are learned from an additional layer of interpretation through generated descriptions.

Overall, these comprehensive results reinforce our main findings: while both models exhibit consistency across modalities, MoralCLIP's moral specialization enables recognition of abstract ethical relationships that transcend surface-level similarities. The trade-offs observed---lower cross-modal similarity scores but stronger moral coherence---highlight a critical challenge in adapting multimodal systems to morally grounded tasks: balancing literal semantic correspondence with the need for higher-level moral abstraction. 

\subsection{Additional Retrieval Examples}
\begin{figure*}
    \centering
    \includegraphics[width=1.0\linewidth]{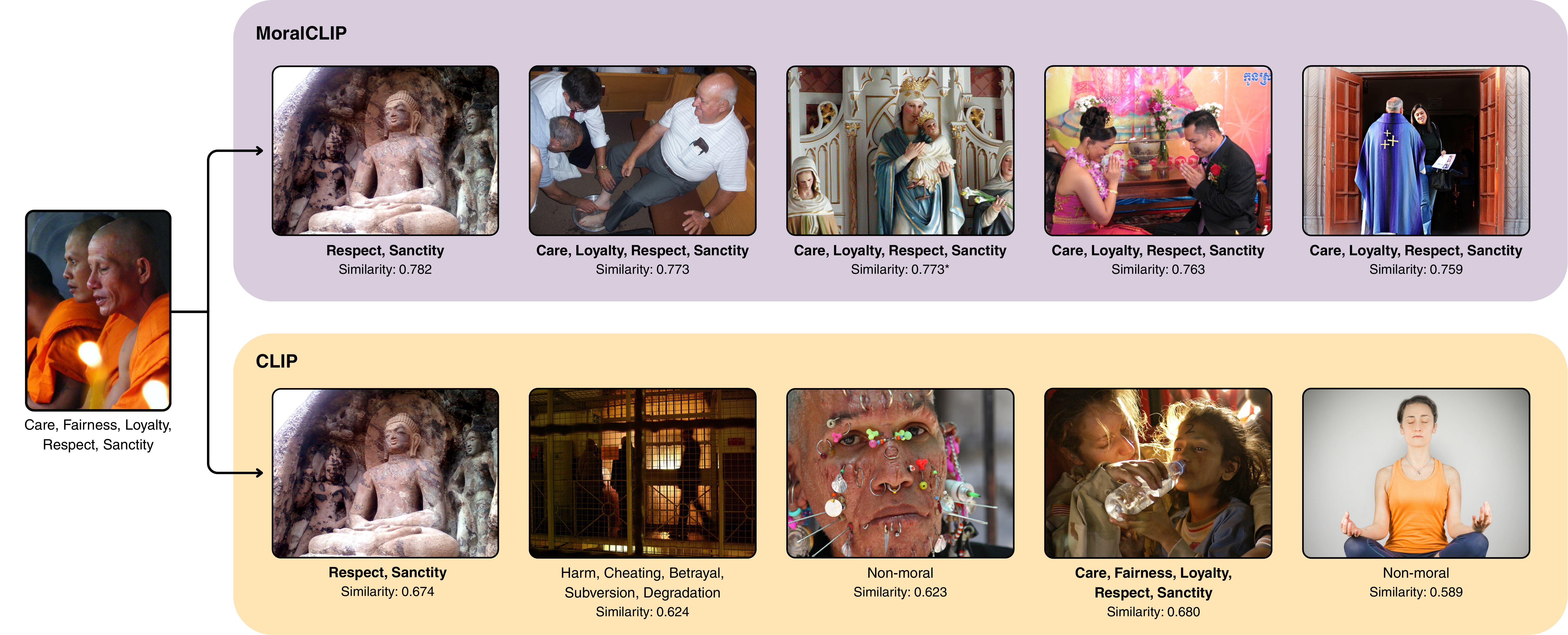}
    \caption{Image-to-Image retrieval comparison between MoralCLIP and CLIP models on the test set. Given a query image depicting a person in a religious setting, MoralCLIP retrieves images emphasizing themes of respect, religion, and human dignity across diverse contexts, while CLIP extracts content with similar colors and low-level visual characteristics. Similarity scores represent cosine similarity. The moral labels in bold match the query's label.}
 \label{fig:extra_I2I}
\end{figure*}

\begin{figure*}
    \centering
    \includegraphics[width=1.0\linewidth]{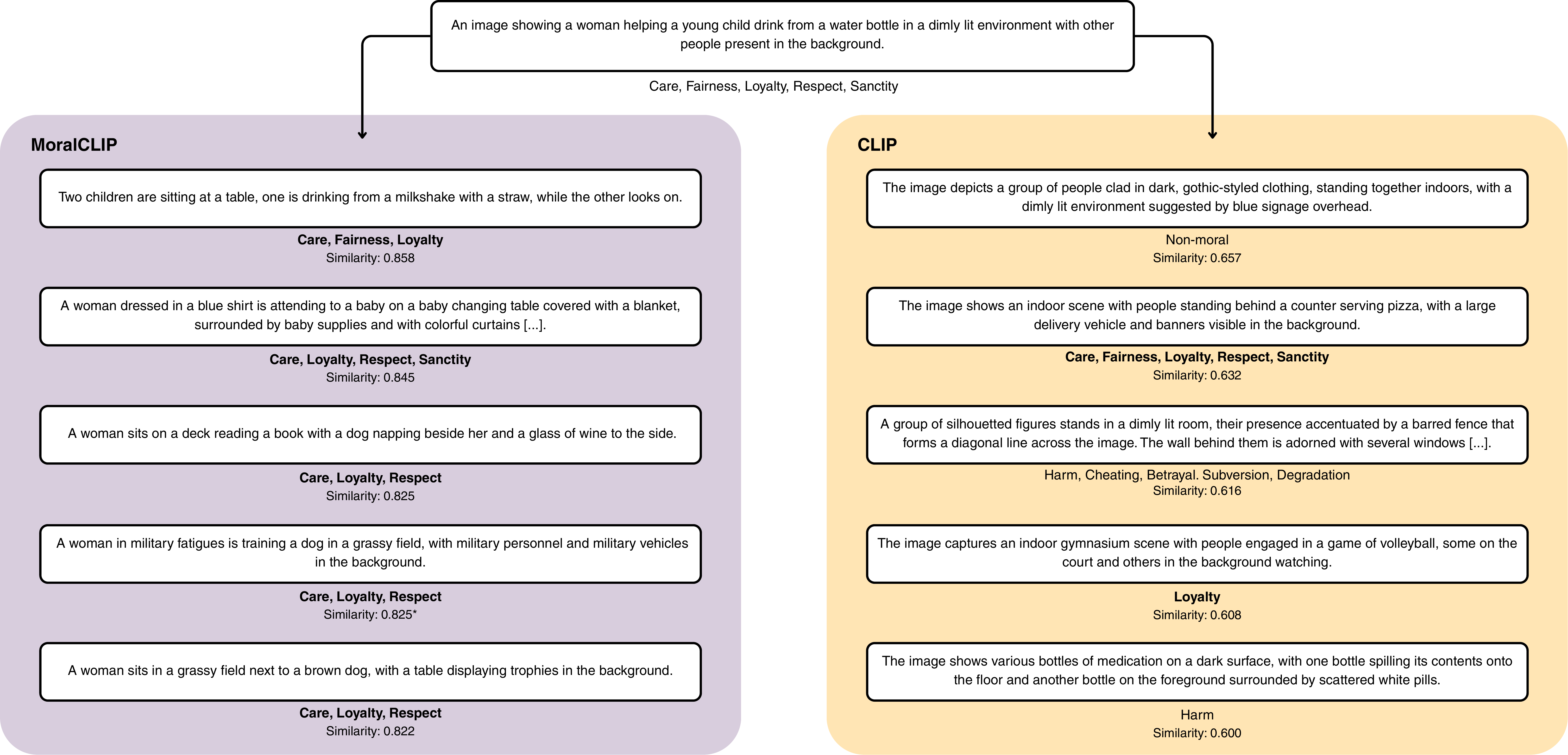}
    \caption{Text-to-Text retrieval comparison between MoralCLIP and CLIP models on the test set. Given a query describing a woman helping a child with care and assistance, MoralCLIP demonstrates remarkable consistency in retrieving descriptions that emphasize nurturing relationships across diverse contexts: from childcare and pet training to simple companionship scenarios. In contrast, CLIP extracts more semantically varied content that includes both positive and negative settings. Similarity scores represent cosine similarity. The moral labels in bold match the query's label.}
    \label{fig:extra_T2T}
\end{figure*}

\begin{figure*}
    \centering
    \includegraphics[width=1.0\linewidth]{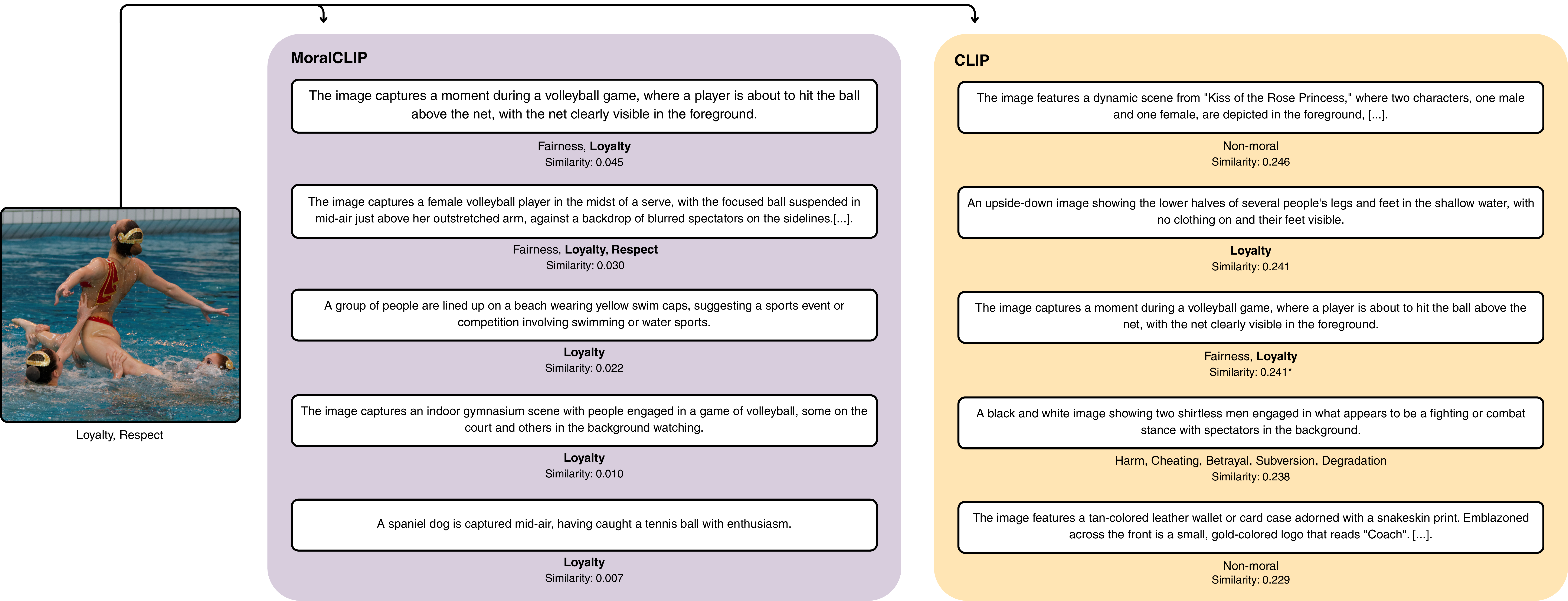}
    \caption{Image-to-Text retrieval comparison between MoralCLIP and CLIP models on the test set. Given an image of synchronized swimming, MoralCLIP mostly retrieves descriptions related to sports and other recreational activities associated with virtuous values. The model showcases semantic coherence by focusing on sports-related activities while maintaining consistent positive moral interpretations. Conversely, CLIP captures a wider variety of content, suggesting that although the model was able to detect the sporting theme, its performance was still heavily influenced by the dominant characteristics of the query image. Similarity scores represent cosine similarity. The moral labels in bold match the query's label.}
    \label{fig:extra_I2T}
\end{figure*}

\begin{figure*}
    \centering
    \includegraphics[width=1.0\linewidth]{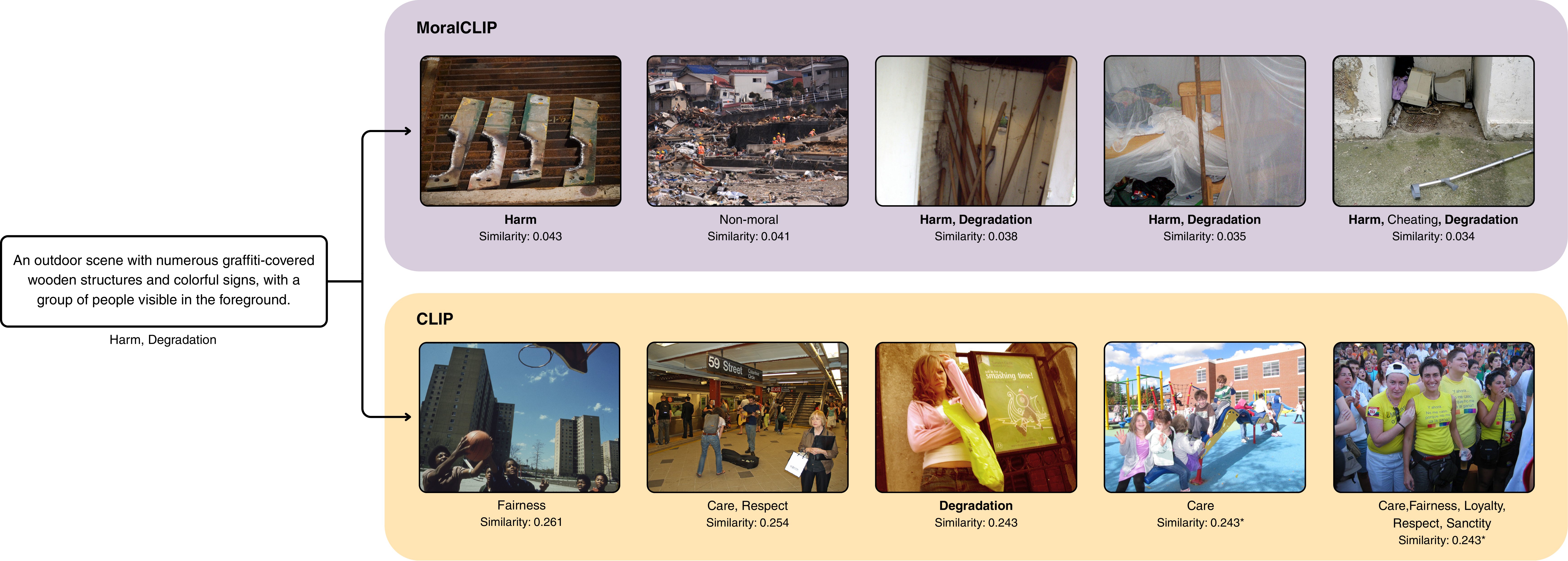}
    \caption{Text-to-Image retrieval comparison between MoralCLIP and CLIP models on the test set. Given a textual query describing urban decay, MoralCLIP consistently retrieves images with negative moral dimensions, and often within the topic of destruction and contamination. Conversely, CLIP extracts content related to both moral poles. Similarity scores represent cosine similarity. The moral labels in bold match the query's label.}
    \label{fig:extra_T2I}
\end{figure*}

Figures~\ref{fig:extra_I2I} to \ref{fig:extra_T2I} present additional examples across all four modality combinations that further support our main findings. These examples demonstrate consistent patterns observed in our analysis: MoralCLIP preserves moral coherence by retrieving content with overlapping moral foundations regardless of semantic content, while CLIP, lacking explicit moral understanding, focuses on semantic and surface-level similarities without regard for moral consistency. The examples span both positive moral dimensions (\textit{Care}, \textit{Respect}, \textit{Loyalty}) and negative ones (\textit{Harm}, \textit{Degradation}), illustrating that MoralCLIP's moral alignment operates effectively across the full spectrum of moral content.

\end{document}